\documentclass[10pt,twocolumn,letterpaper]{article}

\usepackage[pagenumbers]{cvpr}

\usepackage{caption}

\usepackage{colortbl}
\usepackage{enumitem}
\usepackage{algorithm}
\usepackage{algpseudocode}

\DeclareMathOperator*{\argmax}{arg\,max}
\newcommand{\ours}{RATTPO\xspace}
\newcommand{\best}[1]{{\boldmath\textbf{#1}}}
\newcommand{\second}[1]{\underline{#1}}
\newcommand{\suppref}[1]{Supp.~\ref{#1}}
\definecolor{oursrow}{RGB}{252,236,238}
\renewcommand{\paragraph}[1]{\vspace{.35em}\noindent\textbf{#1}\ }

\algrenewcommand\algorithmicrequire{\textbf{Input:}}
\algrenewcommand\algorithmicensure{\textbf{Output:}}

\definecolor{cvprblue}{rgb}{0.21,0.49,0.74}
\usepackage[pagebackref,breaklinks,colorlinks,allcolors=cvprblue]{hyperref}

\def\confName{CVPR}
\def\confYear{2027}

\title{Text-to-Image Prompt Optimization for Unseen Rewards}

\author{Semin Kim \quad Yeonwoo Cha \quad Jaehoon Yoo \quad Seunghoon Hong\\
KAIST\\
{\tt\small \{seminkim, ckdusdn03, wogns98, seunghoon.hong\}@kaist.ac.kr}
}

\begin{document}
\twocolumn[{%
\renewcommand\twocolumn[1][]{#1}%
\maketitle
\vspace{-38pt}%
\begin{center}
\captionsetup{type=figure}
\vspace{-2pt}
\includegraphics[width=\linewidth]{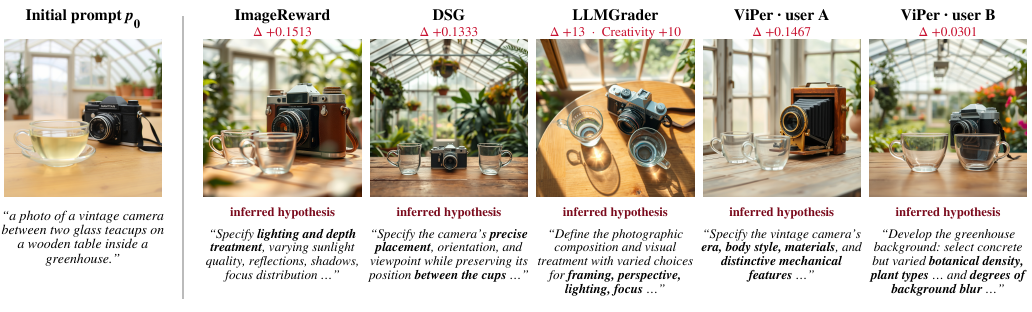}
\vspace{-0.8cm}
\captionof{figure}{\textbf{The right rewrite depends on the reward: from the same prompt, RATTPO infers a different hypothesis about what each unseen reward favors.}
Below each result is the reward hypothesis under which its prompt was proposed.
All rewards share the same initial noise and the same procedure, without any reward-specific adjustment like manual prompting or hyperparameter tuning.
}
\label{fig:teaser}
\end{center}
}]
\begin{abstract}

Text-to-image (T2I) prompt optimization rewrites user prompts to improve the quality of generated images under a target reward.
Existing methods typically assume a known reward, relying on reward-specific training or handcrafted instructions.
We instead study prompt optimization for unseen rewards, which are unknown at design time and accessible only through scores at test time.
We observe that removing reward-specific guidance substantially reduces the search efficiency of LLM-based optimizers, even though their optimization histories contain evidence of what the reward favors.
To address this challenge, we introduce RATTPO (Reward-Adaptive Test-Time Prompt Optimization), a reward-generic method that turns this implicit evidence into explicit search guidance.
At each round, RATTPO contrasts high- and low-scoring prompts to infer competing hypotheses about promising rewriting directions, then generates and evaluates candidates under each hypothesis.
Across four reward families, RATTPO consistently outperforms reward-generic baselines at equal budgets and achieves performance comparable to reward-tailored optimizers without any reward-specific adjustment.

\end{abstract}

\begin{figure*}[t]
    \centering
    \includegraphics[width=1.00\textwidth,clip=false]{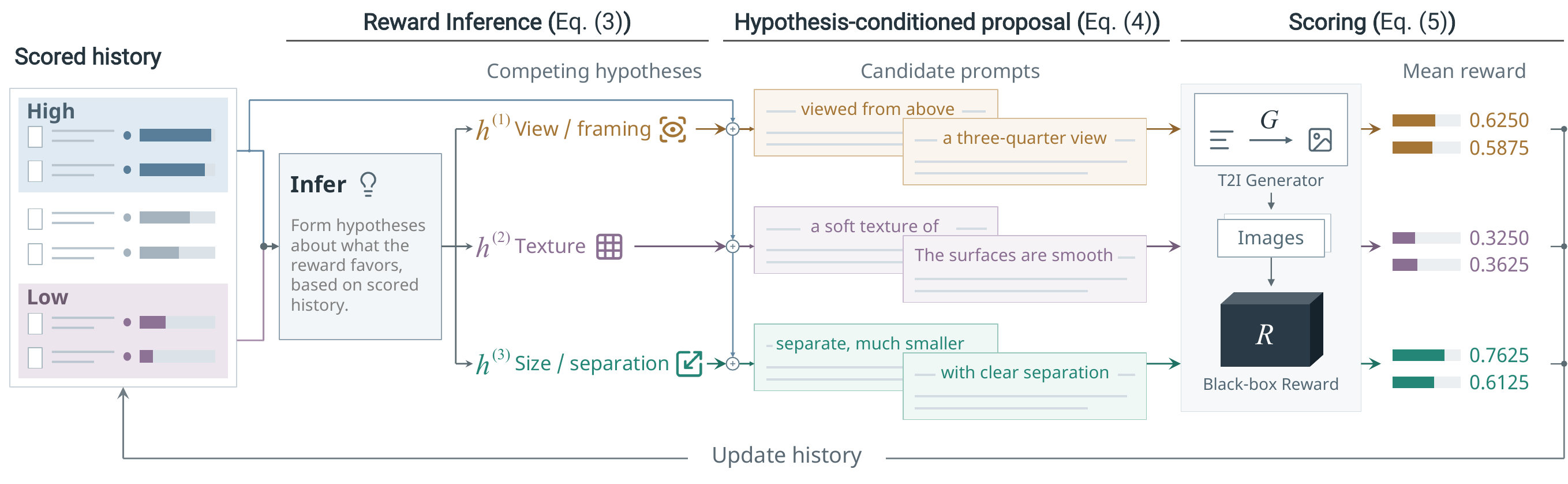}
    \vspace{-1.7em}
    \caption{%
    \textbf{Overview of RATTPO.}
    Each optimization round first infers competing hypotheses about what the target reward favors (\eg, view, texture, or size) by contrasting high- and low-scoring prompts in the optimization history.
    RATTPO then proposes rewrites of the initial prompt $p_0$ that follow each hypothesis, with the high-scoring prompts given as context.
    The rewrites are scored by the black-box target reward $R$ on images from the T2I generator $G$, and the resulting prompt-score pairs are added to the history for the next round.}
    \vspace{-0.6em}
    \label{fig:overview}
\end{figure*}

\vspace{-8pt}
\section{Introduction}
\vspace{-3pt}
Recent advancements in text-to-image (T2I) diffusion models have enabled the generation of high-quality and diverse images from user prompts~\cite{rombach2022high-ldm, ramesh2022hierarchical-dalle2, podellsdxl, BlackForestLabs2024FLUX1}.
Despite this progress, their outputs remain highly sensitive to prompt wording~\cite{liu2022design, oppenlaender2024taxonomy, diab2022stable}, and even subtle prompt changes that preserve the underlying semantics can lead to large differences in generation quality~\cite{mahajan2024prompting, cao2023beautifulprompt}.
To better align generations with user intent, practitioners often iteratively refine the initial prompt through trial and error, a process commonly referred to as prompt engineering.

Prompt engineering seeks prompts that produce \emph{better} images, where what counts as \emph{better} defines the target reward.
In practice, reward models may capture human preference~\cite{xu2023imagereward, schuhmann_improved_aesthetic_predictor,kirstain2023pick-pickscore}, text-to-image consistency~\cite{radford2021learning-CLIP, huang2025t2i, hu2023tifa, cho2024davidsonian-dsg}, or broader holistic criteria and their combinations~\cite{hao2023optimizing-promptist, ma2025inference-noisescale, huang2025t2i, sun2024dreamsync, chen2023x-iqe}.
A reward may also be personalized, representing the preference of an individual user~\cite{salehi2024viper, bai2026personalizetesttimelearning, lee2026personalized}.
Accordingly, several works~\cite{yun2025learning-pag, hao2023optimizing-promptist, mo2024dynamic-pae, manas2024improving-opt2i, wang2024discrete-diffusiondpo} have proposed prompt optimizers that automatically rewrite user prompts into higher-reward prompts.

In realistic deployment, however, the target reward is often specified only at test time.
A single deployed T2I system may need to serve different users with different preferences and evaluation criteria.
Moreover, reward models are continually developed and improved, so the target objective is not a design-time constant.
We call such rewards \emph{unseen rewards}, since they are unknown at design time and accessible only through their scores at test time.
Existing prompt optimizers typically assume a fixed reward setup and overlook this reality.
They are \emph{reward-tailored}, as they rely on either reward-specific training~\cite{yun2025learning-pag, hao2023optimizing-promptist, mo2024dynamic-pae} or handcrafted reward descriptions to guide test-time optimization~\cite{manas2024improving-opt2i}.
As a consequence, they transfer poorly to unseen rewards, which limits their general applicability.
This motivates a \emph{reward-generic} prompt optimizer that can adapt online to diverse reward functions without reward-specific training or handcrafted reward descriptions.

However, we observe that an existing LLM-based prompt optimizer~\cite{yanglarge-opro} suffers from low search efficiency when it is naively adapted to be reward-generic by removing a handcrafted reward description.
The naive adaptation requires roughly twice the prompt budget to reach the same reward (\cref{sec:observation}), even though the scored prompts in its history already carry evidence of what the reward favors.

To address this challenge, we introduce Reward-Adaptive Test-Time Prompt Optimization (RATTPO), a simple yet effective reward-generic prompt optimizer for unseen rewards.
At each round, RATTPO infers competing hypotheses about what the reward favors by contrasting high- and low-scoring prompts in the optimization history (\cref{fig:overview}).
It then proposes prompt rewrites that follow each hypothesis, and the resulting scores serve as evidence for the next round.
These hypotheses play the role of reward-specific guidance, but are induced on the fly from past prompt-reward pairs rather than written by a human.
As shown in \cref{fig:teaser}, the inferred hypotheses can differ across target rewards for the same initial prompt.
The resulting procedure is training-free and gradient-free, making it applicable to diverse reward models, including non-differentiable ones.

\paragraph{Contributions.}
First, we study T2I prompt optimization in the challenging setting of unseen rewards and formulate the requirement of a reward-generic prompt optimizer.
Second, we observe that the search efficiency of an LLM-based prompt optimizer depends on its handcrafted reward description.
Third, we propose RATTPO, which recovers the missing reward knowledge at test time by inferring competing hypotheses about the reward from the scored history.
Lastly, we evaluate RATTPO on four reward families, including per-user preference rewards.
Across all settings, RATTPO consistently improves the initial prompt and achieves the highest reward among the reward-generic optimizers, exceeding the final reward of every reward-generic baseline with only half of the prompt budget.
Without any reward-specific adjustment, it also matches reward-tailored optimizers with handcrafted reward descriptions, including OPT2I~\cite{manas2024improving-opt2i} on its target consistency reward.

\section{Background}
\label{sec:prelim}

\vspace{-3pt}
\paragraph{Prompt optimization for text-to-image generation.}
Text-to-Image (T2I) prompt optimization aims to improve the images generated from a user-provided prompt by rewriting the prompt while preserving its original semantics.
Given an initial prompt $p_0$, a T2I generator $G$ that maps a prompt $p$ and noise $\epsilon$ to an image, and a target reward model $R$, the objective is to find a rewritten prompt that maximizes the expected reward of the generated images:
\begin{equation} \label{eq:prompt_engineering}
    p^\star = \argmax_{p\in \mathcal{M}(p_0)} \mathbb{E}_{\epsilon}\big[R\big(G(p,\epsilon), p_0\big)\big],
\end{equation}
where $\mathcal{M}(p_0)$ denotes the set of prompts that preserve the semantics of $p_0$.
Importantly, the reward is evaluated with respect to the original prompt $p_0$, rather than the rewritten prompt $p$, so that optimization remains grounded in the user's original intent.
In practice, the expected reward is approximated using a fixed set of $n$ noise samples, \ie, $J(p;p_0)=\frac{1}{n}\sum_{i=1}^{n} R(G(p,\epsilon_i),p_0)$.

Optimizing \cref{eq:prompt_engineering} is challenging because the prompt forms a large and discrete search space $\mathcal{M}(p_0)$, and the optimal rewriting strategy depends on the target reward.
Existing approaches address this problem either by training a prompt rewriter for a predefined reward~\cite{hao2023optimizing-promptist, yun2025learning-pag} or by iteratively searching for better prompts using a large language model (LLM) at test time~\cite{yanglarge-opro, manas2024improving-opt2i}.
In this paper, we focus on the latter, which does not require training a separate optimizer for each reward and provides a flexible foundation for adapting to new objectives.

\paragraph{LLM-based prompt optimizers.}
Based on strong in-context learning capability, LLMs can serve as general-purpose black-box optimizers by iteratively generating candidate solutions with previous solutions and their scores as context~\cite{yanglarge-opro, agrawal2026gepa, manas2024improving-opt2i}.
For prompt optimization, a single round of optimization can be written as:
\begin{equation}
\begin{aligned}
  \mathcal{P}_t &= \mathrm{Propose}\big(p_0,\,C_{\mathrm{prop}}(\mathcal{H}_t),\,d_R\big),\\
  \mathcal{H}_{t+1} &= \mathcal{H}_t\cup\big\{(p,J(p; p_0)): p\in\mathcal{P}_t\big\},
\end{aligned}
\label{eq:round}
\end{equation}
where $\mathcal{H}_t$ is the history of evaluated prompt-reward pairs, $C_{\mathrm{prop}}$ selects the relevant examples from the history (\eg, the highest-scoring prompts), and $\mathrm{Propose}$ is an LLM call that generates a set of candidate rewrites $\mathcal{P}_t$.

A key source of guidance in existing LLM-based prompt optimizers is the reward description $d_R$, which specifies what the target reward measures and how prompts should be improved accordingly.
For example, a text-to-image consistency reward may benefit from instructions emphasizing the accurate rendering of entities and attributes, whereas a human-preference reward may favor different visual qualities.
Such reward-specific guidance is intended to help the optimizer focus its search on promising rewriting strategies.
However, providing $d_R$ requires prior knowledge of the target reward and manually designed instructions.

\section{Prompt Optimization for Unseen Rewards}
\label{sec:unseen}
\vspace{-3pt}

\subsection{Problem setup}
\label{sec:problem}
\vspace{-2pt}

\paragraph{Unseen rewards.}
Our goal is to solve the T2I prompt optimization problem in \cref{eq:prompt_engineering} under the setup of \emph{unseen rewards}.
In this setup, the target criteria for judging an improvement are unknown at design time, reflecting realistic deployment scenarios.
Since these criteria can take various forms (\eg, a user's binary like/dislike vote, or multi-aspect feedback from a multimodal LLM), we assume that the criteria are in general represented by a reward model $R$ that outputs a single aggregated scalar.
We further assume that $R$ is a black box, so its gradients are unavailable and only its output can be observed at test time.

\paragraph{Reward-generic prompt optimizer.}
To support unseen rewards, a prompt optimizer should be \emph{reward-generic}, \ie, one that avoids using information about the target reward at design time and instead adapts to the target reward at test time.
It therefore uses the same procedure, prompts, and hyperparameters for every reward, and can be designed once and later used for an unseen reward directly.
In contrast, existing approaches in prompt optimization are \emph{reward-tailored}, as they either train a language model directly with respect to a known target reward or construct an LLM-based prompt optimizer guided by a handcrafted description of it ($d_R$ in~\cref{eq:round}).
For unseen rewards, it is impossible to apply such reward-specific adjustments at design time.

\subsection{Removing the reward description reduces search efficiency}
\label{sec:observation}
\vspace{-2pt}

\begin{figure}[t]
  \centering
  \includegraphics[width=\linewidth]{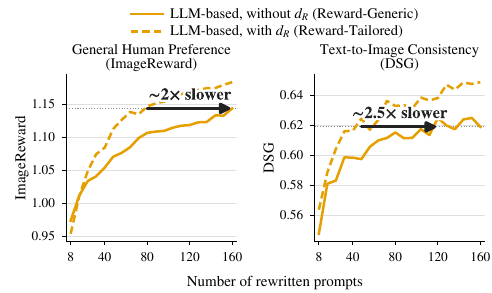}
  \vspace{-1.7em}
  \caption{%
  \textbf{Without the handcrafted reward description $d_R$, the search efficiency of the LLM-based prompt optimizer drops.}
  We plot the reward of the optimized prompt with respect to the prompt budget, \ie, the number of rewritten prompts for the LLM-based prompt optimizer of~\cref{eq:round}, with $d_R$ (reward-tailored, dashed) and without $d_R$ (reward-generic, solid).
  The setup follows the main experiments in \cref{sec:exp}.}
  \label{fig:observation}
  \vspace{-1.4em}
\end{figure}

A reward-generic prompt optimizer is easy to obtain, but making it efficient is hard.
In the LLM-based optimizer of~\cref{eq:round}, information about the target reward comes from both the reward description $d_R$ and the optimization history $\mathcal{H}_t$.
The description states this information explicitly, whereas the history carries it only implicitly through the scores of previous prompts.
One may simply discard $d_R$ and let the optimizer adapt to the target reward solely using the history, which gives a naive reward-generic optimizer.

We empirically observe that the naive optimizer suffers from a large drop in search efficiency.
\Cref{fig:observation} compares the naive optimizer with the reward-tailored optimizer of~\cref{eq:round} at each prompt budget, where the two differ only in the reward description $d_R$ that we handcraft for each target reward (\cref{tab:manual_desc},~\suppref{sec:supp-impl}).
Without $d_R$, it takes roughly twice the prompt budget to achieve a similar target reward.
The gap suggests that the LLM uses reward information more effectively when it is stated explicitly than when it is only implicit in the history, since an explicit description directly restricts the search to a promising subset of $\mathcal{M}(p_0)$.
This motivates converting the implicit information in the history into an explicit description of what the reward favors.

\begin{algorithm}[t]
\caption{\ours}
\label{alg:rattpo}
\small
\begin{algorithmic}[1]
\Require Initial prompt $p_0$, black-box reward $R$,
Prompt budget $b$, candidates per round $m$, number of hypotheses $K$
\Ensure Optimized prompt $\hat p_b$
\State $\mathcal{H}_0\gets\emptyset$
\For{$t=0,\dots,b/m-1$}
  \If{$t=0$}
    \State $\mathcal{P}_0\gets m$ paraphrases of $p_0$ \Comment{no history yet}
  \Else
    \State $\{h_t^{(k)}\}_{k=1}^{K}\gets\mathrm{Infer}\big(p_0,C_{\mathrm{inf}}(\mathcal{H}_t)\big)$
    \ForAll{$k\in\{1,\dots,K\}$ \textbf{in parallel}}
      \State $\mathcal{P}_t^{(k)}\gets\mathrm{Propose}\big(p_0,C_{\mathrm{prop}}(\mathcal{H}_t),h_t^{(k)}\big)$
    \EndFor
    \State $\mathcal{P}_t\gets\bigcup_k\mathcal{P}_t^{(k)}$ \Comment{$m/K$ new candidates each}
  \EndIf
  \State $\mathcal{H}_{t+1}\gets\mathcal{H}_t\cup\{(p,J(p;p_0)) : p\in\mathcal{P}_t\}$
  \State $\hat p_b\gets$ highest-scoring candidate in $\mathcal{H}_{t+1}$
\EndFor
\State \Return $\hat p_b$
\end{algorithmic}
\end{algorithm}

\vspace{-8pt}
\subsection{Reward-Adaptive Test-Time Prompt Optimization}
\label{sec:method}
\vspace{-3pt}

We propose Reward-Adaptive Test-Time Prompt Optimization (RATTPO), a simple yet effective reward-generic prompt optimizer that resolves the search efficiency issue demonstrated in \cref{sec:observation}.
RATTPO infers multiple competing hypotheses about the target reward at test time by contrasting the prompt-score pairs in the optimization history.
Each RATTPO optimization round consists of two stages, followed by the scoring of the proposed prompts (\cref{fig:overview}).

\paragraph{Reward inference.}
The inference stage aims to turn the reward information that is implicit in the optimization history into \emph{explicit} hypotheses that explain why some prompts score higher than others.
Given the initial prompt $p_0$ and an inference context $C_{\mathrm{inf}}(\mathcal{H}_t)$, a single LLM call identifies characteristics associated with higher scores and states them as $K$ different hypotheses $\{h_t^{(k)}\}_{k=1}^{K}$:
\begin{equation}
\label{eq:inference}
  \{h_t^{(k)}\}_{k=1}^{K}
  =\mathrm{Infer}\big(p_0, C_{\mathrm{inf}}(\mathcal{H}_t)\big).
\end{equation}
The inference context $C_{\mathrm{inf}}(\mathcal{H}_t)$ consists of equal numbers of top- and bottom-scoring prompts along with their scores.
We take both ends of the history because such an explanation is easier when high- and low-scoring prompts are contrasted than when only high-scoring prompts are analyzed.

Each hypothesis names one aspect of the depicted image that $p_0$ leaves unspecified or that could be refined (\eg, ``specify the lighting'').
The concrete value of the aspect is left to the proposal stage, which varies it across rewrites so that each hypothesis is explored by several rewrites.
We infer multiple hypotheses because each hypothesis is only a guess about the black-box reward, and several explanations can fit the scores when the history is small.

\paragraph{Hypothesis-conditioned proposal.}
The proposal stage searches for better prompts with each inferred hypothesis in the place of the handcrafted reward description $d_R$ in~\cref{eq:round}.
Every candidate is thus written under an explicit rewriting direction, while the optimizer remains reward-generic.
Specifically, the proposal stage rewrites the initial prompt in $m$ different ways via $K$ LLM calls, each using a different hypothesis to generate $m/K$ candidates:
\begin{align}
\label{eq:proposal}
  \mathcal{P}_t^{(k)}
  &= \mathrm{Propose}\big(p_0,\, C_{\mathrm{prop}}(\mathcal{H}_t), \,h_t^{(k)}\big).
\end{align}
Each call combines the context $C_{\mathrm{prop}}(\mathcal{H}_t)$ built from high-scoring prompts with its hypothesis to propose prompts with higher rewards.
The $K$ calls are independent of each other and can be issued in parallel.

\paragraph{Scoring and iterative adaptation.}
After the two stages, the $m$ generated candidates are used to generate images, which are scored by the target black-box reward:
\begin{equation}
\label{eq:scoring}
  \mathcal{H}_{t+1} =
  \mathcal{H}_t \cup \big\{(p, J(p;p_0)) : p\in\textstyle\bigcup_k\mathcal{P}_t^{(k)}\big\}.
\end{equation}
The resulting prompt-score pairs are stored and used for constructing the contexts in later rounds, so the hypotheses of the next round are inferred from the updated scores.
We repeat the optimization round (\cref{eq:inference,eq:proposal,eq:scoring}) until the given prompt budget $b$ is exhausted, and then return the highest-scoring prompt in the  history.

Compared with the naive reward-generic optimizer in~\cref{sec:observation}, a round of RATTPO adds one inference call and uses $K$ proposal calls instead of one, which increases the cost of a single round.
In return, the inferred hypotheses steer the proposals toward high-reward prompts, so RATTPO reaches the same reward with a smaller prompt budget.
As we show in~\cref{sec:main}, this gain in search efficiency outweighs the additional cost, and RATTPO also reaches the same reward in less wall-clock time (\cref{fig:cost_adjusted_curve}).
\Cref{alg:rattpo} summarizes the full procedure.

\vspace{-3pt}
\section{Related Work}
\label{sec:related-work}
\vspace{-6pt}

\paragraph{Prompt optimization for T2I generation.}
Existing prompt optimizers for T2I generation are tailored to a given reward through either training-based or test-time approaches.
Learning-based methods such as Promptist~\cite{hao2023optimizing-promptist}, PAE~\cite{mo2024dynamic-pae}, and PAG~\cite{yun2025learning-pag} train a prompt rewriter for predefined reward settings.
While effective for trained rewards, these methods require retraining to handle a new reward.
Among test-time methods, OPT2I~\cite{manas2024improving-opt2i} iteratively rewrites prompts with an LLM but relies on handcrafted reward-specific instructions and is evaluated on specific consistency rewards.
PRISM~\cite{he2024automated} targets a specific task of personalized generation from reference images, rather than reward-generic optimization from a user prompt and a scalar reward.
In contrast, we study the setting of unseen rewards defined in \cref{sec:problem}.

\paragraph{LLM-based optimization.}
Recent works have leveraged the strong instruction-following~\cite{ouyang2022training, sanh2022multitask, qin-etal-2024-infobench} and in-context learning capabilities~\cite{brown2020language-icl, wei2022chain-cot, dong2022survey-icl} of LLMs for various optimization tasks~\cite{yanglarge-opro, du2024ipo, manas2024improving-opt2i, he2024automated, zhang2023using, liularge, guo2024connecting-evoprompt}.
Among these, OPRO~\cite{yanglarge-opro} is closely related to our approach, proposing an optimization-by-prompting framework that iteratively queries an LLM using previous optimization history as context.
Other methods guide the LLM with textual feedback on its previous outputs in the form of textual gradients~\cite{pryzant2023automatic, yuksekgonul2024textgrad}, self-feedback~\cite{madaan2023self}, or reflection~\cite{shinn2023reflexion, agrawal2026gepa}.
These methods derive their feedback from a known task and signals such as labeled examples or execution traces.
RATTPO instead considers an objective observed only through scores, and infers competing hypotheses from the scored history.

\section{Experiments}
\label{sec:exp}

\subsection{Setup}
\label{sec:setup}
\vspace{-3pt}

\paragraph{Rewards and prompts.}
We consider four target reward models that cover diverse scenarios, namely human preference (ImageReward~\cite{xu2023imagereward}), text-to-image consistency (DSG~\cite{cho2024davidsonian-dsg}), holistic assessment using a multimodal LLM (LLMGrader~\cite{ma2025inference-noisescale}), and personalized preference (ViPer~\cite{salehi2024viper}).
We use the 64 simple prompts of the ChatGPT evaluation split of~\citet{yun2025learning-pag} for ImageReward and LLMGrader, and a 64-prompt subset of PartiPrompts~\cite{yu2022scaling-parti} used in~\citet{manas2024improving-opt2i} for DSG.
For ViPer~\cite{salehi2024viper}, we construct a personalized reward model for each of 10 users sampled from Pick-a-Pic v2~\cite{kirstain2023pick-pickscore}, and optimize 10 prompts separately for each user.

\paragraph{Baselines.}
We compare RATTPO with both reward-tailored and reward-generic methods.
Promptist~\cite{hao2023optimizing-promptist} and PAG~\cite{yun2025learning-pag} are language models fine-tuned for predefined rewards.
Following PAG, we report the best of 16 beam-search outputs from their official checkpoints.
As a test-time reward-tailored method, we consider OPT2I~\cite{manas2024improving-opt2i}, an LLM-based optimizer as in~\cref{eq:round} specifically tailored for the DSG~\cite{cho2024davidsonian-dsg} reward.
When the target reward differs from the one they are tailored to, these methods are applied without re-training or re-engineering (\ie, in a zero-shot manner).
As a reward-tailored method in the conventional setup where the target reward is known in advance, we consider the LLM-based prompt optimizer of~\cref{eq:round}, which follows OPRO~\cite{yanglarge-opro} and is guided by a reward description $d_R$ handcrafted for each target reward (denoted as OPRO w/ Manual Prompting), and OPT2I on its target reward, DSG.

For reward-generic baselines, we first consider the naive reward-generic optimizer of~\cref{sec:observation}, \ie, the variant of OPRO w/ Manual Prompting that discards $d_R$ (denoted as OPRO).
Additionally, we adapt two other LLM-based optimizers, EvoPrompt~\cite{guo2024connecting-evoprompt} and GEPA~\cite{agrawal2026gepa}, to T2I prompt optimization without any reward-specific information (denoted as EvoPrompt-DE and GEPA, respectively).
We also construct a naive baseline of simply paraphrasing with an LLM and selecting the best one (Paraphrase-BoN).

\paragraph{Implementation.}
We evaluate RATTPO without reward-specific adjustments, using the same hyperparameters and LLM prompts across all setups.
We report the reward of the returned prompt, \ie, the highest-scoring prompt within budget $b$, evaluated on ten held-out noise samples.
We use \texttt{GPT-5.6 Luna} for all LLM components, including baselines, and SDXL-Turbo~\cite{sauer2024adversarial-SDXL-Turbo} as the T2I model.
All test-time methods use a prompt budget of $b{=}160$, with RATTPO running 20 rounds of $m{=}8$ prompts and $K{=}4$ hypotheses.
For ablations and analyses, we focus on two representative rewards, ImageReward and DSG.
Further details on rewards, datasets, and baselines are provided in \suppref{sec:supp-impl}.

\subsection{Main results}
\label{sec:main}
\vspace{-4pt}

\begin{table*}[t]
\centering
\caption{\textbf{\ours is the best reward-generic method for every reward and budget, and matches the reward-tailored methods at $b{=}160$.}
Target reward at budget $b{=}80$ and $160$.
\best{Best}, \second{second}.
$^*$OPT2I optimized for DSG and evaluated on each target reward.
}
\vspace{-0.2cm}
\label{tab:main}
\footnotesize
\setlength{\tabcolsep}{4.3pt}
\begin{tabular}{@{}l cc cc cc cc@{}}
\toprule
& \multicolumn{2}{c}{ImageReward} & \multicolumn{2}{c}{DSG} & \multicolumn{2}{c}{LLMGrader} & \multicolumn{2}{c}{ViPer} \\
& \multicolumn{2}{c}{\scriptsize ChatGPT-style, 64 prompts} & \multicolumn{2}{c}{\scriptsize PartiPrompts, 64 hard prompts} & \multicolumn{2}{c}{\scriptsize ChatGPT-style, 64 prompts} & \multicolumn{2}{c}{\scriptsize Pick-a-Pic v2, 10 users\,$\times$\,10} \\
\cmidrule(lr){2-3}\cmidrule(lr){4-5}\cmidrule(lr){6-7}\cmidrule(l){8-9}
Method & $b{=}80$ & $b{=}160$ & $b{=}80$ & $b{=}160$ & $b{=}80$ & $b{=}160$ & $b{=}80$ & $b{=}160$ \\
\midrule
Initial prompt $p_0$ & 0.813 & 0.813 & 0.475 & 0.475 & 86.0 & 86.0 & 0.497 & 0.497 \\
\midrule
\multicolumn{9}{@{}l}{\textit{Reward-tailored, to another reward}} \\
\quad Promptist~\cite{hao2023optimizing-promptist} ($b=16$) & 1.092 & 1.092 & 0.552 & 0.552 & 91.9 & 91.9 & 0.533 & 0.533 \\
\quad PAG~\cite{yun2025learning-pag} ($b=16$) & 1.018 & 1.018 & 0.508 & 0.508 & 91.8 & 91.8 & 0.567 & 0.567 \\
\quad OPT2I$^*$~\cite{manas2024improving-opt2i} & 0.828 & 0.827 & -- & -- & 86.5 & 86.6 & 0.477 & 0.478 \\
\midrule
\multicolumn{9}{@{}l}{\textit{Reward-tailored, to this reward}} \\
\quad OPT2I~\cite{manas2024improving-opt2i} & -- & -- & 0.627 & 0.643 & -- & -- & -- & -- \\
\quad OPRO w/ Manual Prompting & \second{1.147} & \second{1.184} & \second{0.633} & \second{0.649} & \best{93.5} & \best{93.6} & -- & -- \\
\midrule
\multicolumn{9}{@{}l}{\textit{Reward-generic}} \\
\quad Paraphrase-BoN & 1.052 & 1.062 & 0.621 & 0.617 & 90.1 & 90.2 & 0.584 & 0.592 \\
\quad OPRO~\cite{yanglarge-opro} & 1.107 & 1.144 & 0.615 & 0.619 & 91.3 & 91.8 & \second{0.600} & \second{0.621} \\
\quad EvoPrompt-DE~\cite{guo2024connecting-evoprompt} & 1.090 & 1.113 & 0.605 & 0.614 & 90.9 & 91.5 & 0.592 & 0.608 \\
\quad GEPA~\cite{agrawal2026gepa} & 1.077 & 1.084 & 0.581 & 0.590 & 89.7 & 90.0 & 0.567 & 0.577 \\
\rowcolor{oursrow} \quad \ours & \best{1.156} & \best{1.191} & \best{0.639} & \best{0.650} & \second{92.9} & \best{93.6} & \best{0.658} & \best{0.694} \\
\bottomrule
\end{tabular}
\vspace{-3pt}
\end{table*}

\begin{figure*}[t]
  \centering
  \includegraphics[width=\textwidth]{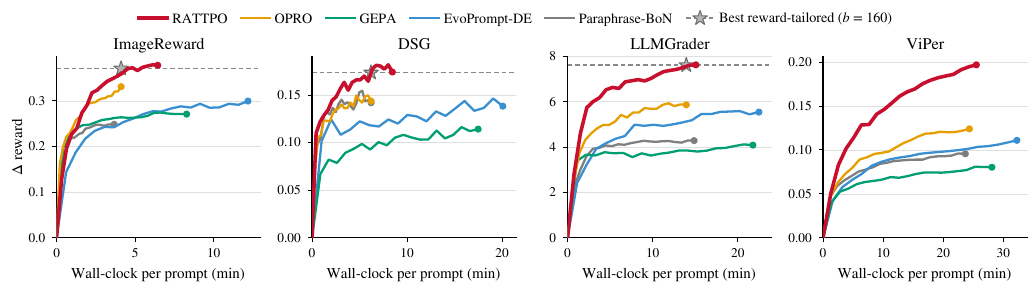}
  \vspace{-1.5em}
  \caption{\textbf{The advantage in search efficiency also holds in wall-clock time.}
  We plot the gain in target reward of the best-so-far prompt over the initial prompt at each round against the wall-clock time per prompt.
  Dots mark $b{=}160$.
  The gray star marks the best reward-tailored result at $b{=}160$ for each reward (OPRO w/ Manual Prompting).
  The reward curves over the prompt budget are in \cref{fig:curves} (\suppref{app:additional}).}
  \vspace{-0.5em}
  \label{fig:cost_adjusted_curve}
\end{figure*}

\begin{figure*}[!t]
  \centering
  \includegraphics[width=\textwidth]{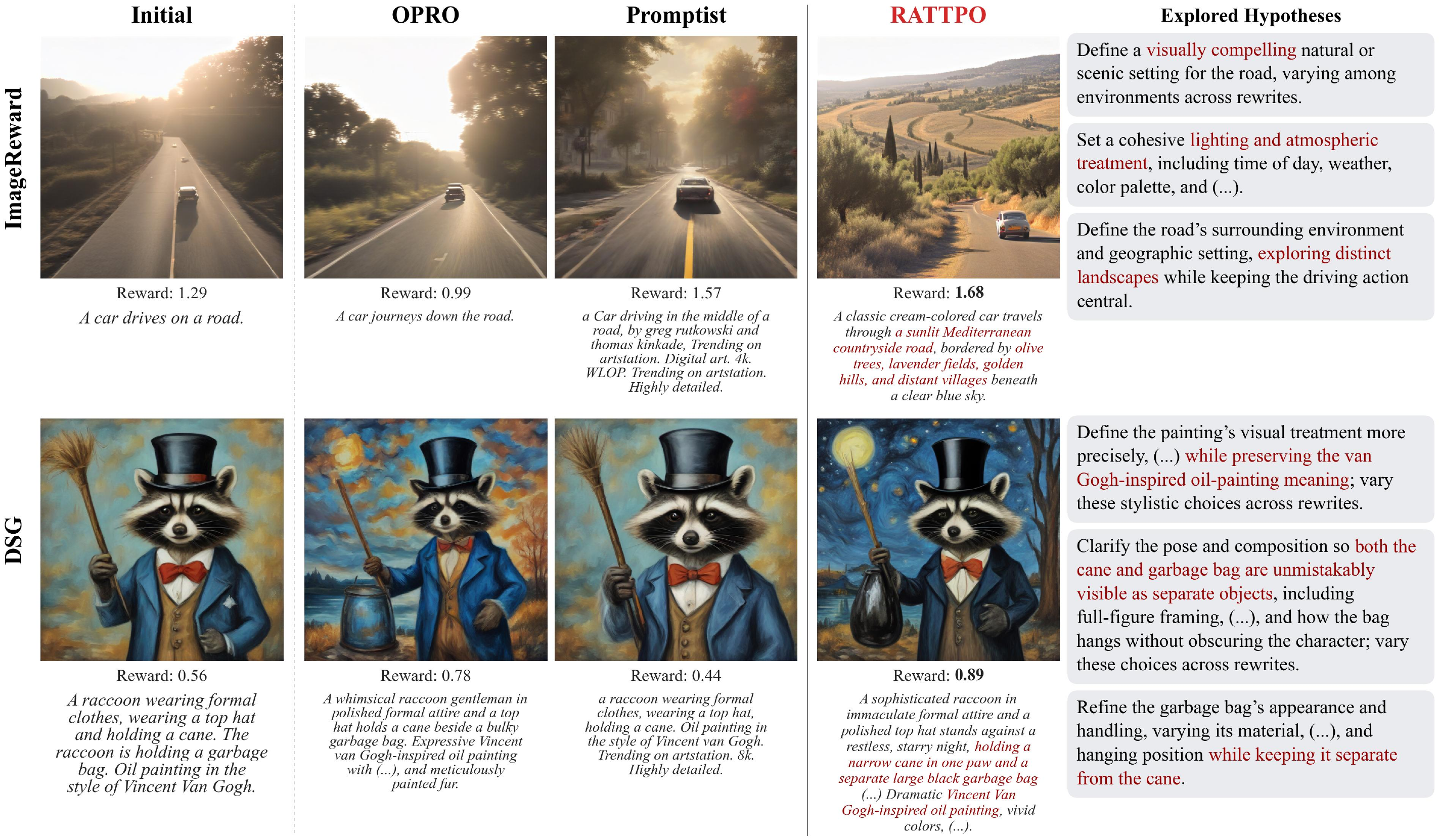}
  \vspace{-1.5em}
  \caption{%
  \textbf{The RATTPO-optimized prompt enhances generation, and the generated hypotheses are reward-specific.}
  For each reward, we show the images generated from the initial prompt and from the prompts optimized by OPRO, Promptist and RATTPO, with the reward and the prompt below each image.
  The right column lists hypotheses that RATTPO used while optimizing the shown prompt, where the phrases that appear to be reflected in the prompt are marked in red. See \suppref{app:additional} for additional results.}
  \label{fig:main_qual}
\end{figure*}

\paragraph{Comparison with reward-generic methods.}
As shown in \cref{tab:main} and qualitatively in \cref{fig:main_qual}, RATTPO consistently improves the initial prompt and achieves the highest reward among the reward-generic methods on all four reward families, at both budgets.
The gain is pronounced on the personalized reward (ViPer), where each user has a distinct reward and a tailored description is difficult to prepare in advance.
As the procedure and hyperparameters are shared across all rewards (\cref{sec:setup}), the results indicate that a single reward-generic optimizer can adapt to diverse unseen rewards.

RATTPO is also the most efficient among the reward-generic methods.
Specifically, RATTPO exceeds the final reward of every reward-generic baseline with only half of the prompt budget on every reward.
This advantage also holds in wall-clock time (\cref{fig:cost_adjusted_curve}), which accounts for the additional LLM calls of RATTPO and is measured with the same GPU type and API concurrency limit for every method.
RATTPO reaches the final reward of every reward-generic baseline in less wall-clock time, on all four rewards.
The savings from a smaller prompt budget thus outweigh the cost of inferring hypotheses.
The efficiency benefit also grows as rewards become more expensive to query, as in the setups motivating our work, such as user feedback or proprietary multimodal LLMs.

\paragraph{Comparison with reward-tailored methods.}
RATTPO is also comparable to the reward-tailored baselines, which are given reward-specific descriptions that are unavailable in the unseen reward setup.
With the full budget, RATTPO matches OPRO w/ Manual Prompting on all three rewards where a manual description is available (gray star in \cref{fig:cost_adjusted_curve}).
On DSG, RATTPO also matches OPT2I, even though the instructions of OPT2I are engineered specifically for this reward.
RATTPO thus recovers the search efficiency lost when the reward description is removed (\cref{sec:observation}), without any manual effort to describe the reward.

In contrast, reward-tailored methods do not transfer reliably when they are applied to a reward that differs from the one they are tailored to.
Promptist and PAG, which are primarily trained for aesthetics, show relatively small improvement under the unseen text-to-image consistency reward (DSG), while showing some gains for general human preference (ImageReward).
On DSG, both fall below every reward-generic method, including the simple Paraphrase-BoN.
OPT2I, optimizing against the DSG, also shows marginal or no gain over the initial prompts for the other rewards.
These results highlight the need for reward-generic prompt optimizers.
We also note that reward-tailored methods require either retraining for every reward at a considerable cost (\eg, 80 V100 GPU days for Promptist~\cite{hao2023optimizing-promptist}) or manually rewriting the reward description, while RATTPO adapts to new rewards without any modification.

\begin{table*}[t]
\centering
\begin{minipage}[t]{0.345\textwidth}
\centering
\caption{%
\textbf{RATTPO shows a focused gain for the target user's preference.}
Gain in the ViPer reward over the initial prompt at $b{=}160$, measured with the reward of the user that the prompts are optimized for (target user) and with the rewards of the other nine users (mean).}
\vspace{-3pt}
\label{tab:viper_cross}
\footnotesize
\setlength{\tabcolsep}{6pt}
\begin{tabular}{@{}lcc@{}}
    \toprule
    Method & Target user & Other users \\
    \midrule
    Paraphrase-BoN & $+0.096$ & $+0.024$ \\
    OPRO & $+0.124$ & $+0.023$ \\
    EvoPrompt-DE & $+0.111$ & $+0.007$ \\
    GEPA & $+0.080$ & $+0.009$ \\
    \rowcolor{oursrow} \ours & $+0.197$ & $+0.021$ \\
    \bottomrule
\end{tabular}
\end{minipage}\hfill
\begin{minipage}[t]{0.33\textwidth}
\centering
\caption{%
\textbf{The gain comes from explicitly inferring what the reward favors.}
We replace the inference stage with alternative forms of feedback while the proposal stage is fixed.
None is identical to OPRO.}
\vspace{-6pt}
\label{tab:abl-inference}
\footnotesize
\setlength{\tabcolsep}{2.5pt}
\begin{tabular}{@{}lcc@{}}
    \toprule
    & ImageReward & DSG \\
    \cmidrule(lr){2-2}\cmidrule(l){3-3}
    Inference stage & $b{=}80$ / $b{=}160$ & $b{=}80$ / $b{=}160$ \\
    \midrule
    None~\cite{yanglarge-opro} & 1.107 / 1.144 & 0.615 / 0.619 \\
    Raw history & 1.104 / 1.141 & 0.627 / 0.632 \\
    Self-Critique~\cite{madaan2023self} & 1.133 / 1.179 & 0.614 / 0.623 \\
    Reflection~\cite{shinn2023reflexion} & 1.129 / 1.165 & 0.622 / 0.630 \\
    \rowcolor{oursrow} \ours & \best{1.156} / \best{1.191} & \best{0.639} / \best{0.650} \\
    \bottomrule
    \end{tabular}
\end{minipage}\hfill
\begin{minipage}[t]{0.265\textwidth}
\centering
\caption{%
\textbf{Using multiple competing hypotheses helps.}
We vary the number of hypotheses per round $K$.
$K=4$ performed the best.}
\label{tab:abl-K}
\footnotesize
\setlength{\tabcolsep}{2.5pt}
\begin{tabular}{@{}lcc@{}}
    \toprule
    & ImageReward & DSG \\
    \cmidrule(lr){2-2}\cmidrule(l){3-3}
    & $b{=}80$ / $b{=}160$ & $b{=}80$ / $b{=}160$ \\
    \midrule
    $K=1$ & 1.091 / 1.124 & 0.614 / 0.622 \\
    $K=2$ & 1.115 / 1.171 & 0.619 / 0.622 \\
    \rowcolor{oursrow} $K=4$ & \best{1.156} / \best{1.191} & \best{0.639} / \best{0.650} \\
    $K=8$ & 1.152 / 1.183 & 0.634 / 0.645 \\
    \bottomrule
    \end{tabular}
\end{minipage}
\vspace{-0.5em}
\end{table*}

\subsection{Analysis}
\label{sec:analyses}
\vspace{-4pt}

We first examine whether the adaptation of RATTPO is specific to the target reward, and then ablate its design choices and check the properties of the optimized prompts.
Unless otherwise noted, the analyses use ImageReward and DSG.

\paragraph{Reward-specific rewriting.}
Qualitative results in \cref{fig:teaser,fig:main_qual} show that the hypotheses generated by RATTPO are reward-specific guidance.
In \cref{fig:teaser}, the hypotheses for the same initial prompt address different aspects depending on the reward.
The hypothesis for ImageReward specifies the lighting and depth of the photo to enhance general image aesthetics, whereas the one for DSG specifies the placement of the camera between the cups to focus on correct scene composition.
The hypotheses also differ between the two users in personalized reward setup, focusing on the design of the camera for one user and on the greenhouse background for the other.
\Cref{fig:main_qual} further shows that these hypotheses are reflected in the optimized prompts.
For ImageReward, the hypotheses ask for a visually compelling scenic setting, and the optimized prompt places the car on ``a sunlit Mediterranean countryside road'' bordered by olive trees and lavender fields.
For DSG, the hypotheses ask to keep the cane and the garbage bag visible as separate objects, and the optimized prompt describes ``a narrow cane in one paw and a separate large black garbage bag''.
These examples show that RATTPO is capable of adapting to the target reward through the reward inference stage, which can be observed qualitatively from the inferred hypotheses.

To quantitatively examine whether the resulting gain is specific to the target reward, \cref{tab:viper_cross} evaluates the prompts optimized for one ViPer user with the rewards of the other nine users.
RATTPO shows the highest gain for the target user, while its gain for the other users remains similar to that of the baselines.
If RATTPO simply wrote generally better prompts, its gain for the other users would also exceed that of the baselines.
The additional gain of RATTPO thus comes from adapting to the target user's preference, instead of finding a general prompt rewriting direction.

\paragraph{Inference stage ablation.}
\Cref{tab:abl-inference} compares the reward inference stage of RATTPO with alternative ways of using the optimization history, where the proposal stage and the budget are kept the same.
\emph{Raw history} gives the inference context to the proposal stage as is, and checks whether showing more of the history is sufficient.
\emph{Self-Critique} is a Self-Refine~\cite{madaan2023self}-style variant that generates critiques of the best prompt of the last optimization round without observing its score, and checks whether feedback without scores is sufficient.
\emph{Reflection} is a Reflexion~\cite{shinn2023reflexion}-style variant that constructs reflections from the prompt-score pairs of the last optimization round and its reflections from the previous three rounds, and checks whether feedback on the recent rounds is sufficient.
Self-Critique and Reflection generate four critiques or reflections per round, which take the place of the $K{=}4$ hypotheses in the proposal calls.
Raw history gives no gain over the variant without inference on ImageReward and only a part of the gain of RATTPO on DSG.
Self-Critique improves ImageReward, while its effect on DSG varies with the budget. Reflection gives a moderate gain on both rewards.
All variants perform worse than RATTPO.
These results indicate that an explicit statement of what the reward favors guides the search better than alternative forms of feedback.

\paragraph{Competing hypotheses.}
\Cref{tab:abl-K} varies the number of hypotheses $K$ while the number of prompts per round is fixed, so each hypothesis receives $m/K$ rewrites.
We find that using multiple hypotheses is beneficial, with the reward saturating at the default $K{=}4$.
If we use a single hypothesis, all proposals of a round commit to a single guess about the black-box reward, which narrows the search.
With multiple hypotheses, however, each round can explore several competing directions, providing more evidence about the target reward for the next round.

\begin{table}[t]
\centering
\caption{%
\textbf{The proposal and the inference stages require different history contexts.}
The history context of the proposal call and of the inference call is built from the best prompts only (Best) or from equal numbers of the best and the worst prompts (Mixed).}
\vspace{-0.6em}
\label{tab:abl-history-context}
\footnotesize
\setlength{\tabcolsep}{2.5pt}
\begin{tabular}{@{}llcc@{}}
    \toprule
    \multicolumn{2}{@{}l}{History context} & ImageReward & DSG \\
    \cmidrule(r){1-2}\cmidrule(lr){3-3}\cmidrule(l){4-4}
    Proposal & Inference & $b{=}80$ / $b{=}160$ & $b{=}80$ / $b{=}160$ \\
    \midrule
    \rowcolor{oursrow} Best & Mixed & \best{1.156} / \best{1.191} & \best{0.639} / \best{0.650} \\
    Best & Best & 1.131 / 1.183 & 0.624 / 0.639 \\
    Mixed & Mixed & 1.132 / 1.174 & 0.613 / 0.632 \\
    \bottomrule
    \end{tabular}
\vspace{-0.8em}
\end{table}

\paragraph{History construction strategy.}
RATTPO builds the contexts of its two stages differently.
The proposal call is given only the best prompts in the history, as is common in LLM optimizers~\cite{yanglarge-opro, manas2024improving-opt2i, guo2024connecting-evoprompt}, while the inference call is given both the best and the worst ones (Mixed).
\Cref{tab:abl-history-context} shows that both choices matter.
When the inference context is also built from the best prompts only, the reward drops on both rewards.
This supports the contrastive inference context motivated in \cref{sec:method}.
Conversely, when the proposal context is built from both the best and the worst prompts, the reward drops as well.
The two stages benefit from different contexts because the inference stage has to explain why some prompts score higher than others, whereas the proposal stage has to write better prompts, for which high-scoring prompts are useful examples.

\paragraph{Semantic preservation.}
Recall that the prompt optimization problem in \cref{eq:prompt_engineering} requires preserving the meaning of the user prompt, which is implemented by instructing the LLM in RATTPO.
To examine whether RATTPO adheres to this restriction, we report how often an LLM judge finds that an optimized prompt drifts from the initial prompt in its style, entities, attributes, counts or relations.
The judge finds no sign of drift for RATTPO (0.0\% for every element on both rewards), indicating that RATTPO does not obtain its gain by changing what the prompt depicts.
Note that the same judge finds drift in the prompts of the trained rewriter (Promptist), most notably on DSG (7.8\% to 15.6\% depending on the element), which agrees with our manual inspection.
The results of all methods are in \suppref{app:additional}.

\paragraph{Held-out rewards.}
We examine whether the gain of RATTPO carries over to rewards not used during optimization.
We evaluate the images generated from the returned prompts with a held-out reward that is correlated with the target reward, \ie, HPSv2.1~\cite{wu2023human} for the prompts optimized for ImageReward, and DSG with a different VQA model for those optimized for DSG.
The returned prompts show a positive gain over the initial prompts for each held-out reward, from 0.260 to 0.279 on HPSv2.1 and from 0.504 to 0.635 on DSG with the different VQA model.
The gain in the target reward thus transfers to rewards that measure a similar criterion.
Full results are in \suppref{app:additional}.

\paragraph{LLM and T2I generality.}
Lastly, we examine whether RATTPO generalizes to other LLMs and T2I models.
We replace the default LLM with Llama 4 Maverick or Mistral Small 4, and separately replace the default T2I model with FLUX.1-schnell~\cite{BlackForestLabs2024FLUX1}.
As shown in \suppref{app:additional}, RATTPO outperforms Paraphrase-BoN and OPRO in every configuration, showing that it is robust to the choice of the LLM and T2I backbones.

\vspace{-3pt}
\section{Conclusion}
\label{sec:conclusion}
\vspace{-4pt}
We studied text-to-image prompt optimization for unseen rewards, accessible only through their scores at test time.
We observed that removing handcrafted reward descriptions substantially reduces the search efficiency of LLM-based optimizers, despite the reward information implicit in their optimization histories.
To address this challenge, we proposed RATTPO, a reward-generic optimizer that infers competing hypotheses from the scored history.
Across four reward families, RATTPO outperforms existing reward-generic methods and
performs comparably to reward-tailored methods without reward-specific adjustments.
Our analyses further confirm the benefits of explicit reward inference and competing hypotheses.

\clearpage
\section*{LLM Usage Statement}
LLMs were used as an integral component of the proposed method for rewriting user-given T2I prompts, as detailed in \cref{sec:method}.
We used Claude Code (Opus 5.5) and Codex (GPT-6 Astra) to assist with language editing, manuscript refinement, and the development of research ideas.
Generative AI tools were also used to assist in preparing related work, generating and debugging code, designing and analyzing experiments, preparing figures, and creating or annotating data.
AI-generated materials were reviewed and revised by the authors, who take full responsibility for the scientific content and reported results.

{
    \small
    \bibliographystyle{ieeenat_fullname}
    \bibliography{main}
}

\clearpage
\setcounter{page}{1}
\maketitlesupplementary

\setcounter{section}{0}
\renewcommand{\theHsection}{supp.\Alph{section}}
\renewcommand{\thesection}{\Alph{section}}

\section{Implementation Details}
\label{sec:supp-impl}

\paragraph{Sampling from the T2I model.}
We use SDXL-Turbo~\cite{sauer2024adversarial-SDXL-Turbo} as the default T2I model.
Images are sampled in half precision (\texttt{fp16}).
Furthermore, for the T2I model generality experiment, we use FLUX.1-schnell~\cite{BlackForestLabs2024FLUX1} in \texttt{bf16} precision.
SDXL-Turbo and FLUX.1-schnell are timestep-distilled models, and thus we use their default samplers with the number of sampling steps set to one and four, respectively.
Images are generated at a resolution of $512{\times}512$ for SDXL-Turbo and $1024{\times}1024$ for FLUX.1-schnell, without classifier-free guidance.
When calculating rewards, we estimate the expectation in~\cref{eq:prompt_engineering} by averaging rewards over generated images from ten different initial diffusion noise samples.
The noise samples are shared across methods and rounds.
The rewards reported in the paper are computed on ten other noise samples that are not used during the optimization, which are also shared across methods.
Specifically, the prompt returned at each prompt budget is selected with the noise samples used for the optimization, and is then evaluated on the held-out noise samples.

\paragraph{Hyperparameters.}
For RATTPO, we perform 20 rounds where each round proposes 8 prompts.
We select the 8 best prompts from the history for the context of the proposal stage, and the 16 best and 16 worst prompts for the inference stage.
The inference stage generates $K{=}4$ hypotheses, and thus each hypothesis is used to propose two prompts per round.
As the history is empty in the first round, the first round paraphrases the initial prompt without using hypotheses (\cref{alg:rattpo}).
We query \texttt{GPT-5.6 Luna} with its reasoning disabled.
The prompts and the hypotheses are obtained as a JSON array of strings by using the structured output of the LLM API.
Llama 4 Maverick and Mistral Small 4 in the LLM generality experiment are queried through OpenRouter.
For baselines, we use the default setup if possible.
To match the prompt budget (\ie, 160 generated prompts), we adjust the number of iterations of OPT2I~\cite{manas2024improving-opt2i}, resulting in 20 optimization iterations where each iteration evaluates 8 prompts, with the initial prompt counted as one of the 8 prompts of the first iteration.
For DSG, whose maximum value is 1, all test-time methods stop the optimization once a prompt attains the maximum reward, and the returned prompt is used for the remaining budgets.

\paragraph{Reward models.}
In our experiments, we consider various rewards that can be categorized into (1) human preference, (2) text-to-image consistency, (3) holistic assessment using a multimodal LLM (MLLM), and (4) personalized preference.
Details about the reward models are provided in \cref{tab:reward_models}.
We employ widely adopted reward models for each category, namely ImageReward~\cite{xu2023imagereward}, DSG~\cite{cho2024davidsonian-dsg}, LLMGrader~\cite{ma2025inference-noisescale}, and ViPer~\cite{salehi2024viper}.
For LLMGrader, we follow the setup in~\citet{ma2025inference-noisescale}. The prompt used for querying the LLM is shown in \cref{fig:llmgrader-prompt}.
We use \texttt{Gemini 3.1 Flash-Lite} as the multimodal LLM of LLMGrader, which is given the generated image and the initial prompt.
For DSG, we use \texttt{GPT-6 Luna} both to generate the questions from the initial prompt and to answer them as the visual question answering (VQA) model.
For ViPer, we use the proxy metric model released by~\citet{salehi2024viper}, which predicts the probability that a user prefers a given image, conditioned on the images that the user preferred and did not prefer.

\begin{table}[t]
\centering
\footnotesize
\setlength{\tabcolsep}{4pt}
\caption{\textbf{Reward models used in our experiments.}
Each reward model is listed with its category and a short description.}
\renewcommand{\arraystretch}{1.12}
    \begin{tabular}{p{0.2\linewidth} p{0.2\linewidth} p{0.47\linewidth}}
    \toprule
    \textbf{Category} & \textbf{Reward Model} & \textbf{Description} \\

    \midrule
    Human \par Preference
    & ImageReward \par \cite{xu2023imagereward}
    & Trained on human preference annotations to capture general human preferences. \\

    \midrule
    Text-to-Image \par Consistency
    & DSG~\cite{cho2024davidsonian-dsg}
    & Generates atomic questions with a dependency graph from the prompt, then scores images using a visual question answering model. \\

    \midrule
    Holistic MLLM \par Assessment
    & LLMGrader \par  \cite{ma2025inference-noisescale}
    & Provides an overall score and five sub-scores: prompt accuracy, creativity, visual quality, consistency, and emotional or thematic resonance. We use the overall score as the optimization target. \\

    \midrule
    Personalized \par Preference
    & ViPer \par \cite{salehi2024viper}
    & Predicts how likely an individual user is to prefer an image, given a set of images that the user preferred and did not prefer. \\

    \bottomrule
    \end{tabular}
\label{tab:reward_models}
\vspace{-1em}
\end{table}

\begin{figure*}[t]
\centering
\fbox{\begin{minipage}{0.96\textwidth}
        \footnotesize
        You are a multimodal large-language model tasked with evaluating images generated by a text-to-image model.
        Your goal is to assess each generated image based on specific aspects and provide a detailed critique, along with a scoring system.
        The final output should be formatted as a JSON object containing individual scores for each aspect and an overall score.
        Below is a comprehensive guide to follow in your evaluation process:

        \smallskip
        1. Key Evaluation Aspects and Scoring Criteria:
        For each aspect, provide a score from 0 to 100, where 0 represents poor performance and 100 represents excellent performance.
        For each score, include a short explanation or justification (1-2 sentences) explaining why that score was given.
        The aspects to evaluate are as follows:

        a) Accuracy to Prompt

        Assess how well the image matches the description given in the prompt.
        Consider whether all requested elements are present and if the scene, objects, and setting align accurately with the text.
        Score: 0 (no alignment) to 100 (perfect match to prompt).

        b) Creativity and Originality

        Evaluate the uniqueness and creativity of the generated image.
        Does the model present an imaginative or aesthetically engaging interpretation of the prompt?
        Is there any evidence of creativity beyond a literal interpretation?
        Score: 0 (lacks creativity) to 100 (highly creative and original).

        c) Visual Quality and Realism

        Assess the overall visual quality, including resolution, detail, and realism.
        Look for coherence in lighting, shading, and perspective.
        Even if the image is stylized or abstract, judge whether the visual elements are well-rendered and visually appealing.
        Score: 0 (poor quality) to 100 (high-quality and realistic).

        d) Consistency and Cohesion

        Check for internal consistency within the image.
        Are all elements cohesive and aligned with the prompt?
        For instance, does the perspective make sense, and do objects fit naturally within the scene without visual anomalies?
        Score: 0 (inconsistent) to 100 (fully cohesive and consistent).

        e) Emotional or Thematic Resonance

        Evaluate how well the image evokes the intended emotional or thematic tone of the prompt.
        For example, if the prompt is meant to be serene, does the image convey calmness?
        If it’s adventurous, does it evoke excitement?
        Score: 0 (no resonance) to 100 (strong resonance with the prompt’s theme).

        \smallskip
        2. Overall Score

        After scoring each aspect individually, provide an overall score, representing the model’s general performance on this image.
        This should be a weighted average based on the importance of each aspect to the prompt or an average of all aspects.Now grade the image based on the above criteria.
        Below is the prompt:

        Prompt: \{\texttt{prompt}\}
\end{minipage}}
\caption{\textbf{LLM prompt of the LLMGrader reward.}
We follow the prompt used in~\citet{ma2025inference-noisescale}.}
\label{fig:llmgrader-prompt}
\end{figure*}

\paragraph{Datasets.}
For the ImageReward and LLMGrader experiments, we directly use the ChatGPT evaluation split used in PAG~\cite{yun2025learning-pag}.
The dataset consists of 64 prompts, which are generated by querying ChatGPT for brief sentences that describe images in around five words (\eg, ``A bird sits on a branch.'').
For the PartiPrompts~\cite{yu2022scaling-parti} dataset used for the DSG~\cite{cho2024davidsonian-dsg} reward setup, we prepare the dataset following OPT2I~\cite{manas2024improving-opt2i}.
Specifically, we use the first 50 prompts from each of four categories (``Properties and Positioning'', ``Quantity'', ``Fine-grained Detail'', ``Complex'').
As the category ``Properties and Positioning'' contains only 35 prompts, the resulting dataset has 185 initial prompts in total.
As the DSG score easily saturates, we select hard prompts among them.
Specifically, we use the 64 prompts with the lowest DSG scores of the initial prompt, where the score of each prompt is averaged over 10 images generated by SDXL-Turbo.
Lastly, for ViPer, we sample 10 prompts from the validation split of Pick-a-Pic v2~\cite{kirstain2023pick-pickscore} and 10 users with at least 8 valid, non-tied preference annotations from its training split.
For each user, we construct a personalized reward model by conditioning a shared ViPer model on 8 preferred--non-preferred image pairs.
Every prompt is optimized separately for each user, resulting in 100 user-prompt pairs.

\begin{table*}[t]
\centering
\footnotesize
\setlength{\tabcolsep}{6pt}
\caption{\textbf{Handcrafted reward descriptions $d_R$ used for OPRO w/ Manual Prompting.}
Each description is given verbatim.}
\renewcommand{\arraystretch}{1.12}
    \begin{tabular}{p{0.11\linewidth} p{0.84\linewidth}}
    \toprule
    \textbf{Reward Model} & \textbf{Handcrafted reward description $d_R$} \\

    \midrule
    ImageReward
    & ImageReward evaluates human preference along three criteria: **image-text alignment** (the image contains the subjects, attributes, quantities, and relationships specified by the prompt), **fidelity** (objects have coherent, realistic shapes, properties, and interactions without malformed or implausible details), and **harmlessness** (the image avoids pornographic, violent, prejudicial, disturbing, or otherwise harmful content). \\

    \midrule
    DSG
    & DSG measures coverage of facts in the original scene using fixed visual yes/no questions. A failed prerequisite can also invalidate dependent facts, and the reward is averaged across ten images. Use the scored rewrite history to form tentative explanations of which wording makes the original entities, attributes, or relations easier to render. Retain useful wording from strong candidates and test focused changes to an uncertain part of the scene instead of uniformly adding detail to everything. Across candidates, vary which potential bottleneck is addressed and how it is expressed. Preserve every requested fact; additional descriptive or presentation detail is useful only when it helps make those facts visually recognizable. Treat historical score differences as noisy evidence rather than proof that any particular phrase is responsible. \\

    \midrule
    LLMGrader
    & The reward model evaluates generated images based on prompt accuracy, creativity, visual quality, internal consistency, and emotional or thematic resonance. It assesses whether the elements and setting match the prompt while offering an imaginative, aesthetically engaging interpretation. Visual quality is judged by resolution, detail, and lighting, even in stylized or abstract images. The reward model also checks that all components appear natural and are free of unintended distortions. \\

    \bottomrule
    \end{tabular}
\label{tab:manual_desc}
\end{table*}

\paragraph{Baselines.}
We provide some additional details for the baselines used in our experiments.
\begin{itemize}[leftmargin=*]
\item Promptist~\cite{hao2023optimizing-promptist} and PAG~\cite{yun2025learning-pag} generate their responses without querying the target reward.
Following PAG~\cite{yun2025learning-pag}, we generate 16 responses and select the one with the highest reward on the noise samples used for optimization, which corresponds to a prompt budget of $b{=}16$.
\item The OPT2I~\cite{manas2024improving-opt2i} baseline in our experiments is implemented by using the LLM prompt of the original paper.
As in the original paper, its history is initialized with the initial prompt and its score.
The LLM prompt is shown in \cref{fig:metaprompt-baselines}.
We use the same LLM (\texttt{GPT-5.6 Luna}) as RATTPO for a fair comparison.
While the OPT2I paper claims that OPT2I can be used for arbitrary text-to-image consistency scores, it uses a different LLM prompt for each choice of the score.
Among the experimental setups used in our main experiments, the DSG reward on PartiPrompts is the only setup for which the LLM prompt for OPT2I is provided, and thus we include OPT2I as a baseline tailored to the target reward only in that setup.
For the other rewards, OPT2I is applied to the initial prompts of each setup without re-engineering its LLM prompt.
Specifically, the prompts are optimized with DSG as the reward and are then evaluated under the target reward (OPT2I$^*$ in \cref{tab:main}).
\item OPRO is the LLM-based prompt optimizer of~\cref{eq:round} without the reward description $d_R$.
It shares the base template with the proposal stage of RATTPO and uses the 8 best prompts in the history as its context, and thus corresponds to RATTPO without the reward inference stage.
OPRO w/ Manual Prompting additionally includes the handcrafted reward description $d_R$ in the template, where the description used for each reward is listed in \cref{tab:manual_desc}.
The description is appended to the sentence that states the task in \cref{fig:metaprompt-prop}, \ie, right after the initial prompt.
We do not consider OPRO w/ Manual Prompting for ViPer, as ViPer is a personalized reward for which handcrafting a description for every individual user is difficult.
\item EvoPrompt-DE follows the differential evolution variant of EvoPrompt~\cite{guo2024connecting-evoprompt} with a population of 8 prompts.
The initial population is generated by paraphrasing the initial prompt, and each of the following 19 generations creates one child for every prompt in the population with an LLM call, where the child replaces its parent if it obtains a higher reward.
The LLM call that creates a child follows the template of the official implementation, with its output format replaced by our structured output (\cref{fig:metaprompt-baselines}).
\item For GEPA~\cite{agrawal2026gepa}, we use the official implementation and treat the T2I prompt itself as the text to optimize, starting from the initial prompt.
We provide only the scalar reward as its feedback.
The prompt budget counts the initial prompt and the following 159 proposals, where a repeated proposal reuses its cached reward.
We keep its default reflection prompt except for the output format, which is replaced by our structured output, and describe the task to GEPA with the objective and the background shown in \cref{fig:metaprompt-baselines}.
\item Paraphrase-BoN, similar to the paraphrasing baseline used in OPT2I~\cite{manas2024improving-opt2i}, generates 8 paraphrases of the initial prompt at every round, using the base template of OPRO without the history.
\end{itemize}

\begin{figure*}[t]
  \centering
  \includegraphics[width=\textwidth]{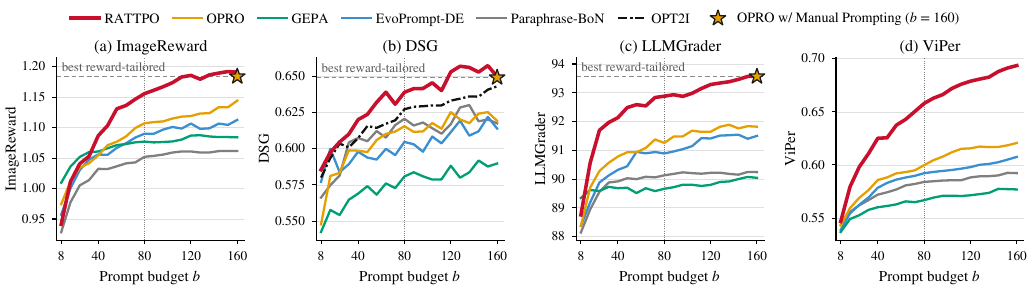}
  \vspace{-1.6em}
  \caption{\textbf{RATTPO shows the best search efficiency among the reward-generic prompt optimizers after the first few rounds, and matches the reward-tailored baselines at $b{=}160$.}  The target reward of the best-so-far prompt with respect to the prompt budget $b$ is plotted.
  Stars mark the $b{=}160$ result of OPRO w/ Manual Prompting, which is tailored to each reward.}
  \label{fig:curves}
\end{figure*}

\begin{figure*}[t]
  \centering
  \includegraphics[width=\textwidth]{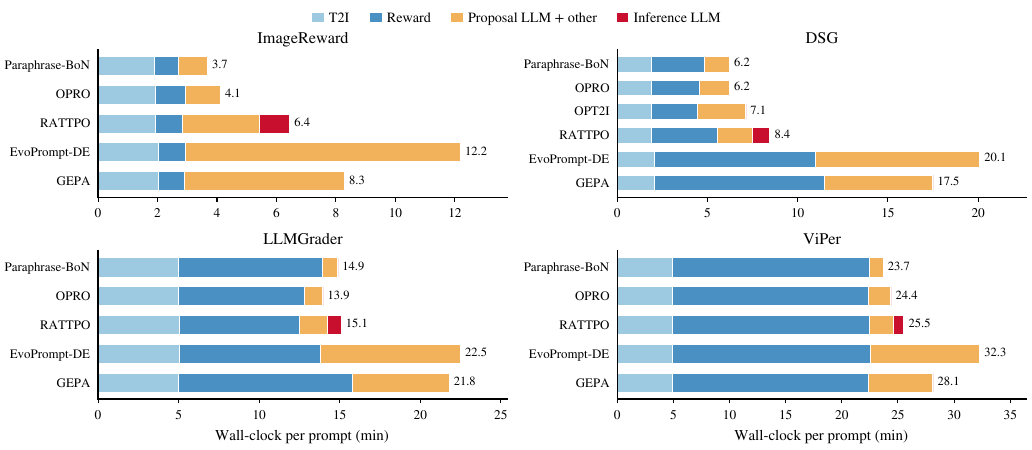}
  \vspace{-1.6em}
  \caption{\textbf{Time per prompt by component on all four rewards.}
  The wall-clock time of a full run ($b{=}160$) for one initial prompt is divided into the image generation, the reward computation, the inference call of RATTPO, and the remaining time.}
  \label{fig:breakdown_full}
\end{figure*}

\paragraph{Wall-clock time.}
For \cref{fig:cost_adjusted_curve,fig:breakdown_full}, we measure the wall-clock time of each optimization round, which includes the LLM calls, the image generation, and the reward computation of the round, and excludes the time for loading the models.
We first average the time of a round over all rounds and initial prompts.
The wall-clock time at a prompt budget $b$ is then this average multiplied by $b/m$, the number of rounds needed to propose $b$ prompts.
Every run uses a single NVIDIA A100 GPU with 40\,GB of memory.
The $K$ proposal calls of RATTPO are issued in parallel, and the reward models that are queried through an API are also queried concurrently for every method.
EvoPrompt-DE and GEPA propose and evaluate one prompt at a time following their original procedures, which results in sequential LLM calls.
In \cref{fig:breakdown_full}, the time is divided into the image generation (T2I), the reward computation (Reward), the inference call of RATTPO (Inference LLM), and the remaining time (Proposal LLM + other).

\paragraph{Cross-user evaluation.}
For \cref{tab:viper_cross}, the images generated from the prompt returned at $b{=}160$ for each user-prompt pair are scored with the personalized reward models of all ten users.
The gain for the target user is computed with the reward model of the user that the prompt is optimized for, and the gain for the other users is the mean over the reward models of the remaining nine users.
Each gain is measured against the reward of the initial prompt under the same reward model, and is averaged over the 100 user-prompt pairs.

\begin{figure*}[t]
\centering
\fbox{\begin{minipage}{0.96\textwidth}
        \footnotesize
        You are evaluating whether an optimized text-to-image prompt preserves the semantics of an initial prompt.

        \smallskip
        Initial prompt: \{\texttt{initial\_prompt}\}\\
        Optimized prompt: \{\texttt{optimized\_prompt}\}

        \smallskip
        Definitions:\\
        * Use ``not\_applicable'' when the initial prompt does not explicitly specify the corresponding constraint. For underspecified prompts such as single adjectives, moods, or aesthetic concepts, evaluate only the constraints that are explicitly present.\\
        * Style drift occurs if a style, medium, rendering type, lighting, mood, atmosphere, aesthetic concept, or visual format explicitly present in the initial prompt is removed or contradicted. Adding compatible style or visual details is not style drift. Use ``not\_applicable'' if the initial prompt does not specify any such constraint.\\
        * Entity drift occurs if a concrete subject, object, character, or scene element explicitly present in the initial prompt is removed or contradicted. Use ``not\_applicable'' if the initial prompt does not specify any concrete entity or scene element.\\
        * Attribute drift occurs if an attribute explicitly specified in the initial prompt is removed or contradicted. Use ``not\_applicable'' if the initial prompt does not explicitly specify any object, subject, or scene attribute.\\
        * Count drift occurs if an explicit number or quantity in the initial prompt is contradicted or changed into an incompatible quantity in the optimized prompt. Do not mark count drift merely because the optimized prompt omits a count word, as long as it preserves the relevant entities and does not imply a different quantity. Use ``not\_applicable'' if the initial prompt does not explicitly specify any number or quantity.\\
        * Relation drift occurs if an explicitly specified relational or interaction constraint is contradicted or changed into an incompatible relation. Relation drift should not be marked merely because the optimized prompt uses different wording, implies the relation through common-sense visual context, or omits the relation while keeping it plausible and not contradicted.\\
        Annotation rule:\\
        Base each label on explicit textual evidence from the initial and optimized prompts. Use ``yes'' only when the optimized prompt contradicts, reverses, or changes an explicitly stated constraint into an incompatible one. Use ``no'' when the constraint is preserved, compatible, implied, or merely underspecified without contradiction. Do not mark drift based on possible unspecified intent.

        \smallskip
        Return only a valid JSON object with the following fields:\\
        \{``style\_drift'': ``yes/no/not\_applicable'', ``entity\_drift'': ``yes/no/not\_applicable'', ``attribute\_drift'': ``yes/no/not\_applicable'', ``count\_drift'': ``yes/no/not\_applicable'', ``relation\_drift'': ``yes/no/not\_applicable'', ``short\_rationale'': ``one sentence''\}
\end{minipage}}
\caption{\textbf{LLM prompt of the judge in the semantic preservation analysis.}
The judge is given an initial prompt and an optimized prompt, and labels the drift in each of the five elements (\cref{tab:semantic_full}).}
\label{fig:semantic-judge-prompt}
\end{figure*}

\paragraph{Semantic preservation.}
For the semantic preservation analysis (\cref{tab:semantic_full}), we use \texttt{Gemini 3.1 Flash-Lite} as the LLM judge.
The judge is given only the initial prompt and the optimized prompt, \ie, the prompt returned at $b{=}160$ ($b{=}16$ for Promptist and PAG), without the name of the method or the reward.
Following the instruction in \cref{fig:semantic-judge-prompt}, the judge labels each of the five elements as yes, no, or not applicable, where the last one is used when the initial prompt does not specify the element.
We report the share of yes labels among the 64 initial prompts of each dataset, and thus a not applicable label is counted as no drift.
OPT2I is evaluated only on DSG, the reward it is tailored to.

\paragraph{Held-out rewards.}
For the held-out reward evaluation (\cref{tab:heldout}), the target and the held-out rewards are computed on the same images, which are generated from the prompts returned at $b{=}160$ with the held-out noise samples.
The returned prompts are selected only with the target reward.
For the prompts optimized for ImageReward, the held-out reward is HPSv2.1~\cite{wu2023human}.
For the prompts optimized for DSG, the held-out reward keeps the questions of DSG and their dependency graph, and only replaces the VQA model that answers the questions, using \texttt{Gemini 3.1 Flash-Lite} instead of \texttt{GPT-6 Luna}.

\section{LLM Prompts}
\label{sec:supp-metaprompt}
\definecolor{burntorange}{rgb}{0.8, 0.33, 0.0}

The LLM prompts used in the proposal stage and in the inference stage can be found in \cref{fig:metaprompt-prop} and \cref{fig:metaprompt-inf}, respectively.
The history and the hypotheses are not available in the first round of RATTPO.
In this case, the LLM prompt for the proposal stage reduces to the prompt for the Paraphrase-BoN baseline, which uses neither the optimization history nor hypotheses.
This corresponds to a prompt that consists of the uncolored text in \cref{fig:metaprompt-prop}.
The LLM prompts of the baselines described in \cref{sec:supp-impl} are shown in \cref{fig:metaprompt-baselines}.

\begin{figure*}[t]
\centering
\fbox{\begin{minipage}{0.96\textwidth}
        \footnotesize
        You are an expert prompt optimizer for text-to-image models.
        Text-to-image models take a text prompt as input and generate images depicting the prompt as output.
        You translate prompts written by humans into better prompts for the text-to-image models.
        Your answers should be concise and effective.

        \smallskip
        Your task is to optimize this initial prompt written by a human: ``\{\texttt{initial\_prompt}\}''.
        \textcolor{blue}{To help with this task, you will also be given 8 previously evaluated prompts.
        Each history prompt has a numerical score; higher scores indicate better prompts.\\
        Use these scores to guide the rewriting process, but do not repeat the history prompts.}\\
        \textcolor{burntorange}{(Hint: \{\texttt{hint}\}.)}

        \smallskip
        \textcolor{blue}{History prompts:\\
        1. \{\texttt{prompt\_1}\} (Score: \{\texttt{score\_1}\})\\
        2. \{\texttt{prompt\_2}\} (Score: \{\texttt{score\_2}\})\\
        (...)\\
        8. \{\texttt{prompt\_8}\} (Score: \{\texttt{score\_8}\})}

        \smallskip
        Now generate 6 paraphrases of the initial prompt which keep the semantic meaning \textcolor{blue}{and that have higher scores than all the prompts above}.
        Preserve the meaning of the original prompt.
        Keep each variation under 70 words.

        \smallskip
        Return 6 unnumbered strings in the JSON array `items'.
\end{minipage}}
\caption{\textbf{LLM prompt of the proposal stage.}
We highlight the \textcolor{blue}{history-related part} and the \textcolor{burntorange}{hypothesis-related part} in colored text. The first paragraph is given as the system prompt.}
\label{fig:metaprompt-prop}

\medskip
\fbox{\begin{minipage}{0.96\textwidth}
        \footnotesize
        You generate diverse improvement hints for black-box text-to-image prompt optimization.

        \smallskip
        Original prompt: ``\{\texttt{initial\_prompt}\}''\\
        The following rewritten prompts have already been evaluated.
        Higher scores are better.\\
        Histories:\\
        Prompt: \{\texttt{prompt\_1}\} (Score: \{\texttt{score\_1}\})\\
        Prompt: \{\texttt{prompt\_2}\} (Score: \{\texttt{score\_2}\})\\
        (...)\\
        Prompt: \{\texttt{prompt\_32}\} (Score: \{\texttt{score\_32}\})\\
        Generate exactly 4 concise hints for improving future rewrites.
        Use this history to identify patterns or characteristics that appear to be associated with higher scores and generate hints for revising the original prompt in ways likely to improve its score.\\
        Each hint must name a distinct aspect of the depicted image that is left unspecified in the original prompt or could be further refined.
        It must instruct the optimizer to make that aspect explicit with concrete details or refinements of its own choosing and to vary those choices across rewrites.
        The hints must be clearly distinct from one another.
        A hint must not prescribe particular details or wording for that aspect; concrete choices are explored and selected automatically.
        Each hint must remain faithful to the original meaning and be independently usable by a prompt optimizer.
        Do not rewrite the original prompt or combine multiple directions into one hint.\\
        Output only 4 numbered hints, one per line.\\
        Return 4 to 4 unnumbered strings in the JSON array `hints'.
\end{minipage}}
\caption{\textbf{LLM prompt of the inference stage.}
The first sentence is given as the system prompt. The history consists of the 16 best and 16 worst prompts.}
\label{fig:metaprompt-inf}
\end{figure*}

\begin{figure*}[t]
\centering
\fbox{\begin{minipage}{0.96\textwidth}
        \footnotesize
        \textbf{OPT2I.}
        You are an expert prompt optimizer for text-to-image models.
        Text-to-image models take a text prompt as input and generate images depicting the prompt as output.
        You translate prompts written by humans into better prompts for the text-to-image models.
        Your answers should be concise and effective.

        \smallskip
        Your task is to optimize this initial prompt written by a human: ``\{\texttt{initial\_prompt}\}''.
        Below are some previous prompts with the consistency of each prompt's visual elements in the generated image via a set of binary questions.
        The prompts are arranged in ascending order based on their overall consistency score, which ranges from 0 to 100 (higher is better).\\
        1. \{\texttt{prompt\_1}\}\\
        overall\_score: \{\texttt{score\_1}\}\\
        evaluation questions:\\
        \{\texttt{question\_1}\} \{\texttt{score\_1\_1}\}\\
        (...)\\
        (...)

        \smallskip
        Generate 8 paraphrases of the initial prompt which keep the semantic meaning and that have higher scores than all the prompts above.
        Focus on optimizing for the visual elements that are not consistent.
        Favor substitutions and reorderings over additions.
        Respond with each new prompt in between \texttt{<PROMPT>} and \texttt{</PROMPT>}, eg:\\
        1. \texttt{<PROMPT>}paraphrase 1\texttt{</PROMPT>}\\
        (...)\\
        8. \texttt{<PROMPT>}paraphrase 8\texttt{</PROMPT>}
\end{minipage}}

\medskip
\fbox{\begin{minipage}{0.96\textwidth}
        \footnotesize
        \textbf{EvoPrompt-DE.}
        Please follow the instruction step-by-step to generate a better prompt.\\
        1. Identify the different parts between the Prompt 1 and Prompt 2:\\
        Prompt 1: \{\texttt{donor\_prompt\_1}\}\\
        Prompt 2: \{\texttt{donor\_prompt\_2}\}\\
        2. Randomly mutate the different parts\\
        3. Combine the different parts with Prompt 3, selectively replace it with the different parts in step2 and generate a new prompt.\\
        Prompt 3: \{\texttt{best\_prompt}\}\\
        4. Crossover the prompt in the step3 with the following basic prompt and generate a final prompt:\\
        Basic Prompt: \{\texttt{parent\_prompt}\}\\
        Return a JSON object with exactly two string fields: ``operations'' and ``final\_prompt''.
        In ``operations'', describe the four operations in order.
        In ``final\_prompt'', provide only the final prompt, without instructions or operation descriptions.
\end{minipage}}

\medskip
\fbox{\begin{minipage}{0.96\textwidth}
        \footnotesize
        \textbf{GEPA (objective).}
        Maximize the returned scalar score by optimizing the actual image prompt, not a reusable rewriting instruction.
        Preserve the meaning of the original prompt.
        Keep the prompt under 70 words.\\
        \textbf{GEPA (background).}
        Original prompt (immutable semantic reference): \{\texttt{initial\_prompt}\}
\end{minipage}}
\caption{\textbf{LLM prompts of the baselines.}
For OPT2I, the history contains up to the 5 best prompts with the scores of their DSG questions. For EvoPrompt-DE, we show the prompt of the LLM call that creates a child, where Prompt 1 and Prompt 2 are randomly selected from the other prompts in the population, Prompt 3 is the best prompt in the population, and the basic prompt is the parent. For GEPA, we show the objective and the background that are given to its official implementation.}
\label{fig:metaprompt-baselines}
\end{figure*}

\twocolumn[{%
  \begin{minipage}{\textwidth}\centering
\captionof{table}{\textbf{Semantic preservation of all compared methods.}
Share (\%) of (\,$p_0$, optimized prompt\,) pairs in which the LLM judge finds a drift in each element (lower is better). The judge uses a different LLM backbone from the prompt optimizers. Promptist and PAG use $b{=}16$, and the other methods use $b{=}160$.}
\label{tab:semantic_full}
\footnotesize
\setlength{\tabcolsep}{5pt}
\begin{tabular}{@{}lcccccccccc@{}}
\toprule
& \multicolumn{5}{c}{Optimized for ImageReward} & \multicolumn{5}{c}{Optimized for DSG} \\
\cmidrule(lr){2-6}\cmidrule(l){7-11}
Method & Style & Entity & Attribute & Count & Relation & Style & Entity & Attribute & Count & Relation \\
\midrule
Promptist & 0.0 & 0.0 & 0.0 & 0.0 & 4.7 & 7.8 & 15.6 & 14.1 & 7.8 & 12.5 \\
PAG & 0.0 & 1.6 & 6.3 & 3.1 & 1.6 & 15.6 & 37.5 & 35.9 & 10.9 & 31.3 \\
\midrule
OPT2I & -- & -- & -- & -- & -- & 0.0 & 1.6 & 1.6 & 0.0 & 1.6 \\
OPRO w/ Manual Prompting & 0.0 & 0.0 & 0.0 & 0.0 & 0.0 & 0.0 & 0.0 & 0.0 & 0.0 & 0.0 \\
\midrule
Paraphrase-BoN & 0.0 & 0.0 & 0.0 & 0.0 & 0.0 & 0.0 & 0.0 & 0.0 & 0.0 & 0.0 \\
OPRO & 0.0 & 0.0 & 0.0 & 0.0 & 0.0 & 0.0 & 1.6 & 0.0 & 0.0 & 0.0 \\
EvoPrompt-DE & 0.0 & 0.0 & 0.0 & 0.0 & 0.0 & 0.0 & 0.0 & 0.0 & 1.6 & 3.1 \\
GEPA & 0.0 & 0.0 & 0.0 & 0.0 & 0.0 & 0.0 & 0.0 & 0.0 & 0.0 & 1.6 \\
\rowcolor{oursrow} \ours & 0.0 & 0.0 & 0.0 & 0.0 & 0.0 & 0.0 & 0.0 & 0.0 & 0.0 & 0.0 \\
\bottomrule
\end{tabular}
  \end{minipage}
  \par\vspace{1.2em}
  \begin{minipage}{\textwidth}\centering
\captionof{table}{%
\textbf{The gain in the target reward transfers to held-out rewards that measure a similar criterion.}
The rewards of $p_0$ and of the RATTPO-optimized prompt at $b{=}160$ are shown, under the target reward (Optimized) and under a held-out reward on the same generated images.
Other VQA changes the VQA model in the DSG reward from \texttt{GPT-6 Luna} to \texttt{Gemini 3.1 Flash-Lite}.
}
\label{tab:heldout}
\footnotesize
\setlength{\tabcolsep}{2.6pt}
\begin{tabular}{@{}lcccc@{}}
\toprule
& \multicolumn{2}{c}{ImageReward} & \multicolumn{2}{c}{DSG} \\
\cmidrule(lr){2-3}\cmidrule(l){4-5}
Method & Optimized & HPSv2.1~\cite{wu2023human} & Optimized & Other VQA \\
\midrule
$p_0$ & 0.813 & 0.260 & 0.475 & 0.504 \\
\rowcolor{oursrow} \ours & 1.191 & 0.279 & 0.650 & 0.635 \\
\bottomrule
\end{tabular}
  \end{minipage}
  \par\vspace{1.2em}
  \begin{minipage}{\textwidth}\centering
\captionof{table}{\textbf{RATTPO can be used with other LLM or T2I backbones.}
The reward gains with respect to the initial prompt are shown for Paraphrase-BoN, OPRO, and RATTPO, under different combinations of LLM and T2I backbones ($b{=}160$).
RATTPO shows the largest reward gain for all combinations.
}
\label{tab:general}
\footnotesize
\setlength{\tabcolsep}{5.2pt}
\begin{tabular}{@{}llcccccc@{}}
\toprule
& & \multicolumn{3}{c}{ImageReward} & \multicolumn{3}{c}{DSG} \\
\cmidrule(lr){3-5}\cmidrule(l){6-8}
LLM & T2I model & Paraphrase-BoN & OPRO & \ours & Paraphrase-BoN & OPRO & \ours \\
\midrule
GPT-5.6 Luna & SDXL-Turbo & $+0.249$ & $+0.331$ & \best{$+0.378$} & $+0.142$ & $+0.144$ & \best{$+0.174$} \\
Llama 4 Maverick & SDXL-Turbo & $+0.240$ & $+0.292$ & \best{$+0.394$} & $+0.147$ & $+0.146$ & \best{$+0.179$} \\
Mistral Small 4 & SDXL-Turbo & $+0.255$ & $+0.359$ & \best{$+0.434$} & $+0.138$ & $+0.148$ & \best{$+0.163$} \\
GPT-5.6 Luna & FLUX.1-schnell & $+0.216$ & $+0.286$ & \best{$+0.396$} & $+0.079$ & $+0.108$ & \best{$+0.131$} \\
\bottomrule
\end{tabular}
  \end{minipage}
  \par\vspace{1.5em}
}]
\section{Additional Results}
\label{app:additional}

This section provides the results that are referred to in the main paper, together with additional qualitative results.

\paragraph{Quantitative results.}
\Cref{fig:curves} plots the target reward of the best-so-far prompt over the prompt budget $b$ on all four rewards, which complements the wall-clock comparison in \cref{fig:cost_adjusted_curve}.
\Cref{fig:breakdown_full} shows the time per prompt by component.
For the analyses in \cref{sec:analyses}, \cref{tab:semantic_full} reports the semantic preservation results of all compared methods, \cref{tab:heldout} reports the held-out reward evaluation, and \cref{tab:general} reports the reward gains under other LLM and T2I backbones.

\begin{figure*}[t]
  \centering
  \includegraphics[width=\textwidth]{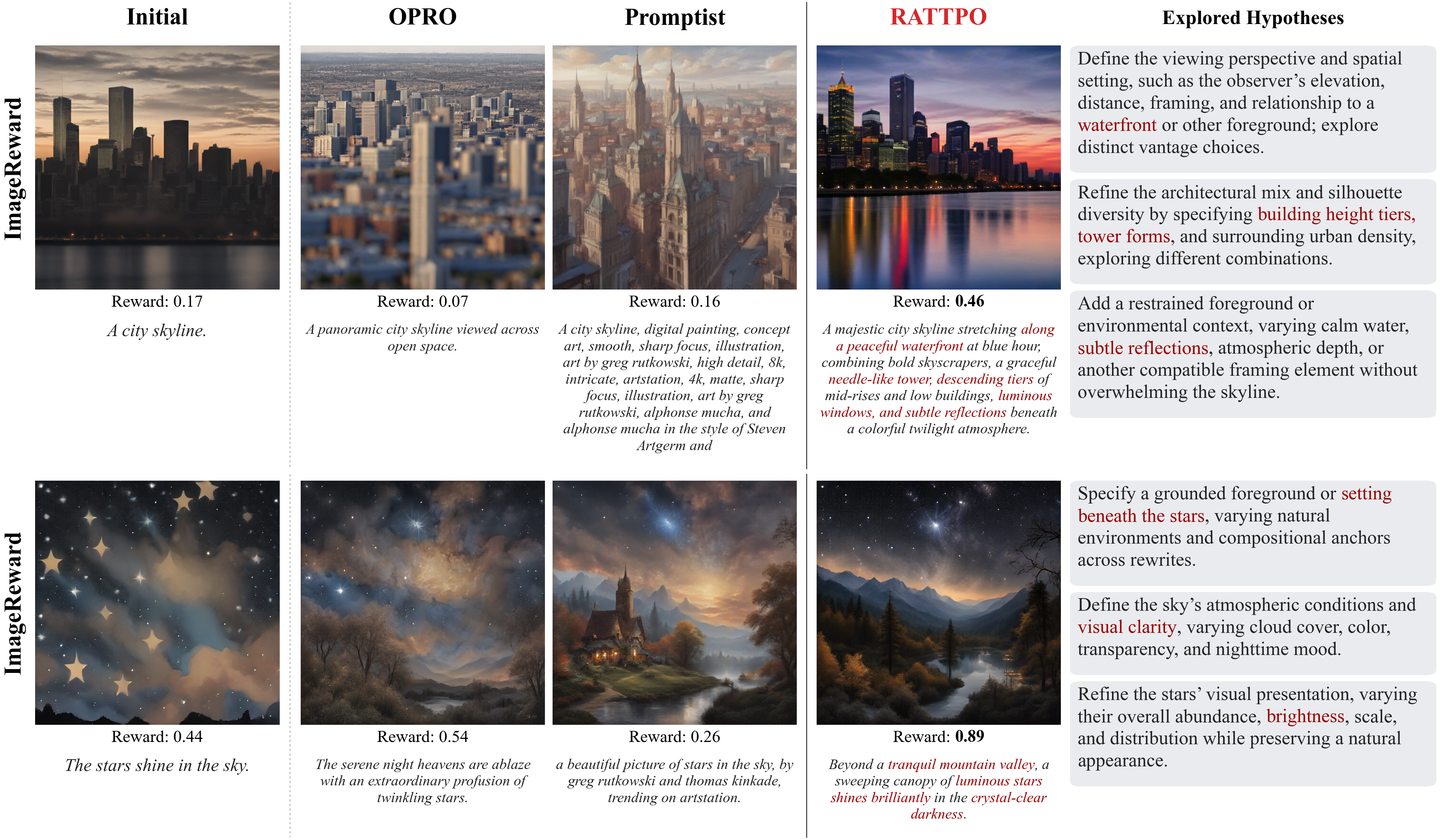}
  \caption{\textbf{Additional qualitative results for ImageReward.}
  As in \cref{fig:main_qual}, we show the images generated from the initial prompt and from the prompts optimized by OPRO, Promptist, and RATTPO, with the reward and the prompt below each image.
  The right column lists hypotheses that RATTPO used while optimizing the shown prompt, where the phrases that appear to be reflected in the prompt are marked in red.
  The images of each example are generated from the same initial noise, and the reward below each image is the reward of that image.}
  \label{fig:supp-qual-imagereward}
\end{figure*}

\begin{figure*}[t]
  \centering
  \includegraphics[width=\textwidth]{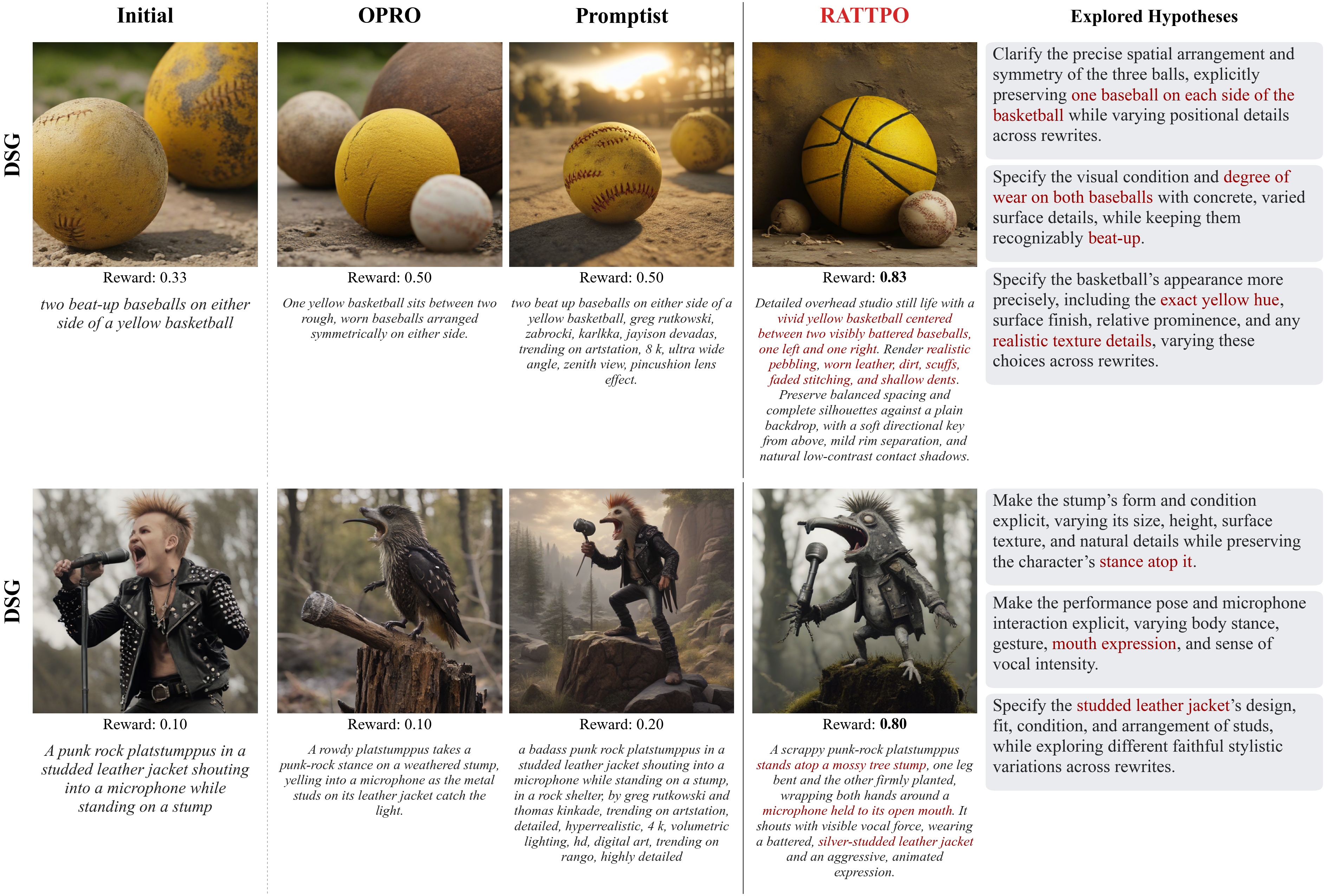}
  \caption{\textbf{Additional qualitative results for DSG.}
  The layout follows \cref{fig:supp-qual-imagereward}.}
  \label{fig:supp-qual-dsg}
\end{figure*}

\begin{figure*}[t]
  \centering
  \includegraphics[width=\textwidth]{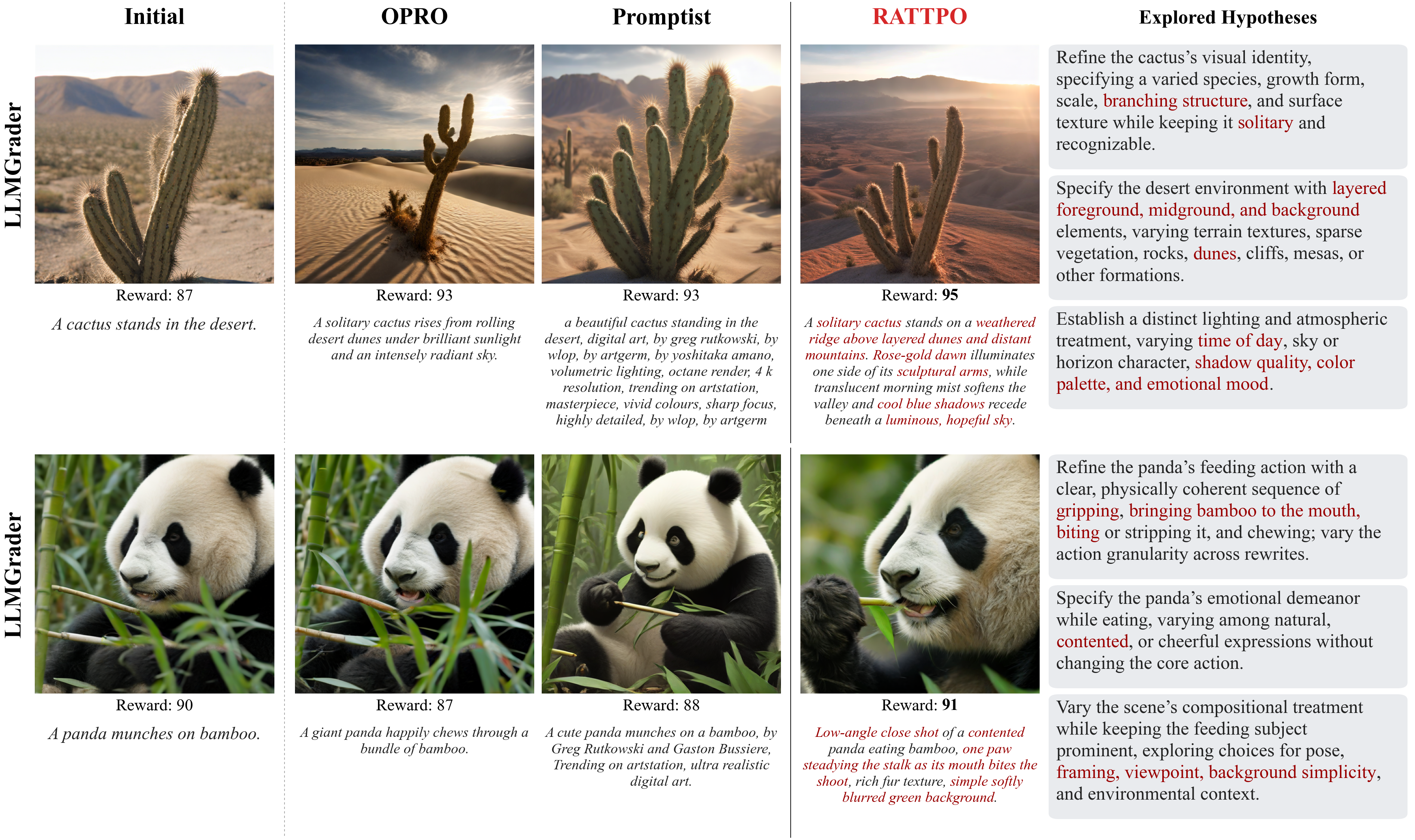}
  \caption{\textbf{Additional qualitative results for LLMGrader.}
  The layout follows \cref{fig:supp-qual-imagereward}.}
  \label{fig:supp-qual-llmgrader}
\end{figure*}

\begin{figure*}[t]
  \centering
  \includegraphics[width=\textwidth]{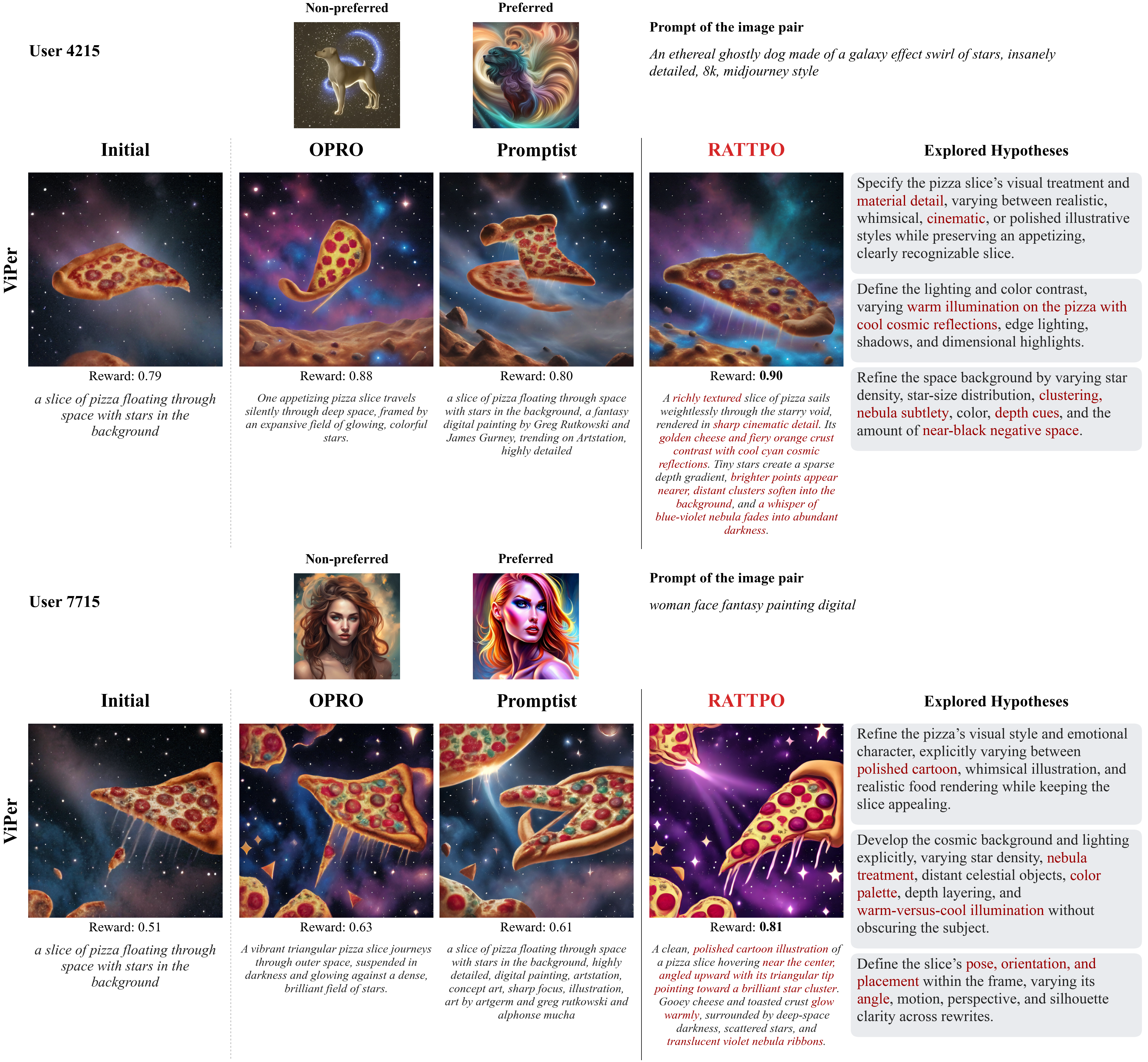}
  \caption{\textbf{Additional qualitative results for ViPer.}
  We show the results of the same initial prompt for two users.
  For each user, the upper row shows one of the preferred--non-preferred image pairs that condition the personalized reward model, along with the prompt of the pair.
  The lower row follows \cref{fig:supp-qual-imagereward}.}
  \label{fig:supp-qual-viper}
\end{figure*}

\paragraph{Qualitative results.}
\Cref{fig:supp-qual-imagereward,fig:supp-qual-dsg,fig:supp-qual-llmgrader,fig:supp-qual-viper} show additional qualitative results for ImageReward, DSG, LLMGrader, and ViPer, respectively.
The figures follow the format of \cref{fig:main_qual}, \ie, each example shows the image, the reward, and the prompt for the initial prompt and for the prompts optimized by OPRO, Promptist, and RATTPO, together with the hypotheses that RATTPO used.
For ViPer, we show the results of the same initial prompt for two users, along with one of the image pairs that condition the personalized reward model of each user.

\end{document}